\documentclass{article}

\usepackage{xcolor}
\usepackage[preprint]{navverse_preprint}

\usepackage{amsmath,amsfonts,amssymb}
\usepackage{booktabs}
\usepackage{caption}
\usepackage{graphicx}
\usepackage{multirow}
\usepackage{subcaption}
\usepackage{listings}
\usepackage{tabularx}
\usepackage{xparse}
\usepackage[most]{tcolorbox}
\usepackage{adjustbox}
\tcbuselibrary{listings,breakable}
\lstdefinestyle{promptstyle}{
  basicstyle=\ttfamily\footnotesize,
  columns=fullflexible,
  breaklines=true,
  breakatwhitespace=true,
  keepspaces=true,
  showstringspaces=false,
  frame=none,
  xleftmargin=0pt,
  aboveskip=0pt,
  belowskip=0pt
}
\usepackage{pifont}
\usepackage{makecell}
\newcommand{\cmark}{\textcolor{teal}{\ding{51}}}
\newcommand{\xmark}{\textcolor{red}{\ding{55}}}

\title{
\makebox[\textwidth][c]{
\begin{minipage}{1.08\textwidth}
\centering
NavVerse: Benchmarking Indoor-to-Outdoor Embodied Navigation in Continuous Robot Simulation
\end{minipage}
}
}

\author{
\bfseries
Junzhe~Wu$^{1}$, Yue~Hu$^{1}$, Zeyu~Han$^{2}$, Po-Hsun~Chang$^{1}$, \\
\bfseries
Yinan~Dong$^{1}$, Behrad~Rabiei$^{1}$, Maani~Ghaffari$^{1}$\\[2pt]
\normalfont
$^{1}$University of Michigan, Ann Arbor 
$^{2}$Tsinghua University\\[3pt]
\href{https://umich-curly.github.io/NavVerse-Benchmark/}
{\texttt{https://umich-curly.github.io/NavVerse-Benchmark/}}
}

\hypersetup{
  pdftitle={NavVerse: Benchmarking Indoor-to-Outdoor Embodied Navigation in Continuous Robot Simulation},
  pdfauthor={Junzhe Wu, Yue Hu, Zeyu Han, Po-Hsun Chang, Yinan Dong, Behrad Rabiei, Maani Ghaffari},
  pdfsubject={Physics-enabled indoor-to-outdoor embodied navigation benchmark},
  pdfkeywords={embodied navigation, indoor-to-outdoor navigation, robot simulation, vision-language navigation}
}

\begin{document}
\maketitle

\begin{figure}[h]
  \centering
  \includegraphics[width=1.0\textwidth]{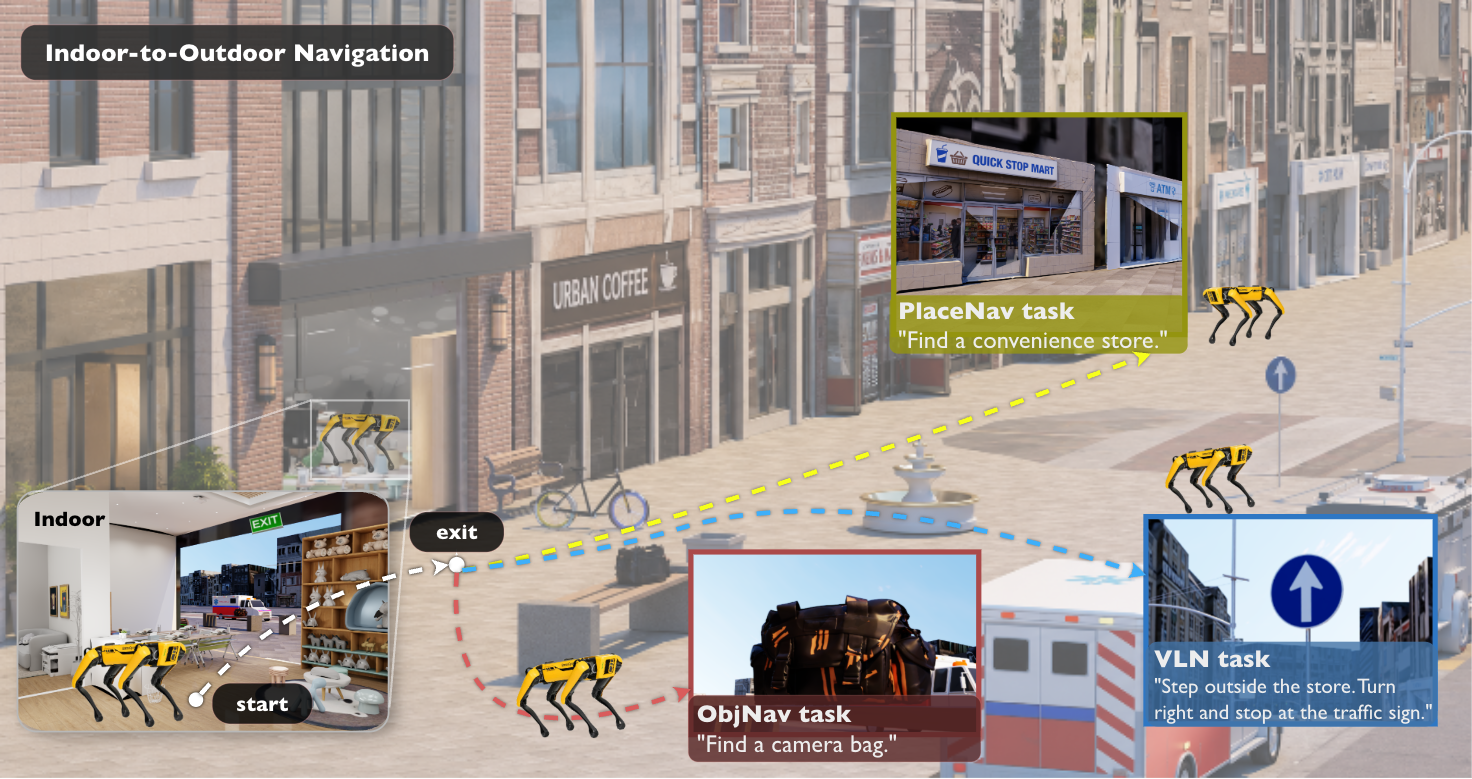}
  \caption{\textsc{NavVerse} connects indoor, outdoor, and hybrid indoor-to-outdoor scenes for ObjNav, VLN, and PlaceNav tasks under continuous physics-enabled robot execution, and evaluates task success, efficiency, and safety.}
  \label{fig:intro}
  \vspace{-2mm}
\end{figure}

\begin{abstract}
Robots deployed in delivery, campus, and emergency-response settings often need to navigate from buildings to streets within a single continuous episode. Existing benchmarks usually evaluate indoor and outdoor navigation separately, and many abstract away robot execution, leaving exit finding, boundary traversal, adaptation, and kinodynamic failures underexplored. We introduce \textsc{NavVerse}, a physics-enabled benchmark for indoor-to-outdoor embodied navigation. \textsc{NavVerse} contains 100 indoor scenes, 50 urban outdoor scenes, and 50 indoor-to-outdoor scenes, and 10,000 episodes spanning Object Navigation, Vision-and-Language Navigation, and Place Navigation tasks, where agents search for semantic points of interest such as restaurants or banks. Agents are evaluated through executable robot interfaces using task-success, path-efficiency, and safety metrics. Zero-shot experiments with RL, VLA, and modular baselines show that current agents remain far from solving cross-context navigation: end-to-end VLAs obtain the highest zero-shot success, while the modular method provides the strongest safety profile. PlaceNav further reveals a clear drop from outdoor to indoor-to-outdoor scenes, indicating that adaptation remains major bottleneck.
\end{abstract}

\section{Introduction}

Navigation is a fundamental prerequisite for robots to physically assist humans in real-world environments.
While recent research has achieved remarkable success in visual navigation within indoor environments~\cite{yin_unigoal_2025,uninavid,li_vln-video_2024,procthor,wei_streamvln_2025,wang_dynam3d_2025,tian_loc4plan_2024,zeng_poliformer_2024,long_instructnav_2024,yokoyama_vlfm_2024,hu_imaginative_2025,yuan_gamap_2024,sun_outdoor_2023}, the deployment of general-purpose mobile robots requires capabilities that extend beyond these structured confines.
Real-world applications, such as last-mile delivery, search and rescue, and campus security, demand agents capable of seamless indoor-to-outdoor transition.
These hybrid scenarios present a transition gap: agents need to discover exits and adapt to scale and appearance shifts while maintaining robust physical interaction with the environment.

Despite this need, existing evaluation platforms are highly limited.
Indoor benchmarks cover increasingly diverse indoor scenes, including scanned scenes in MP3D~\cite{mp3d}/HM3D~\cite{hm3d} and procedural or artist-designed scenes in ProcTHOR~\cite{procthor}/RoboTHOR~\cite{robothor}; however, they remain indoor-only and do not evaluate connected indoor-to-outdoor transitions under continuous robot dynamics.
In contrast, outdoor benchmarks typically rely on offline, non-interactive datasets (e.g., KITTI~\cite{kitti}, nuScenes~\cite{nuscenes}, TouchDown~\cite{chen_touchdown_2020}).
Critically, the transition zone, where agents must reason about tasks such as exiting buildings or traversing curbs, remains underexplored.
Existing benchmarks rarely evaluate these transition behaviors in connected indoor-to-outdoor scenes, where agents must continue acting across the boundary under realistic robot dynamics and traversability constraints.

To bridge this gap, we introduce \textsc{NavVerse}, a unified benchmark designed specifically for indoor-to-outdoor navigation in continuous robot simulation.
Unlike prior works that decouple visual planning from control, \textsc{NavVerse} builds a physics-enabled simulation benchmark that unifies multiple action interfaces, supports different robot embodiments, and provides safety-related diagnostics for full-stack navigation evaluation.
This enables the evaluation of agents not just on path efficiency, but also on physical viability, failure modes, and safety-related metrics across varying terrain conditions.

\textsc{NavVerse} systematizes evaluation across three environment types (indoor, outdoor, and indoor-to-outdoor) and three tasks: Object Navigation (ObjNav), Vision-and-Language Navigation (VLN), and novel Place Navigation (PlaceNav).
Most notably, we introduce PlaceNav as a long-horizon navigation to place-level goals (e.g., locating a nearby restaurant).
This novel yet naturalistic challenge bridges the gap between room-scale and city-scale navigation, requiring agents to transition from indoor spaces to external targets, a routine real-world capability unsupported by existing benchmarks.
Solving PlaceNav requires understanding place-level semantics, grounding urban topology priors, and conducting long-horizon target search under kinodynamic and safety constraints.
Evaluation metrics cover task success, path efficiency, and safety-aware execution.

We conducted extensive evaluations of four representative navigation baselines on \textsc{NavVerse}: an end-to-end visual-language-action (VLA) model, a transformer-based RL policy, a VLA-RL model, and a modular method.
Our results indicate:
i) Cross-context adaptation remains difficult: the end-to-end model performs best among zero-shot baselines, while all evaluated baselines struggle in hybrid indoor-to-outdoor scenarios that require exit finding, boundary crossing, and post-transition adaptation;
and ii) A realistic navigation benchmark must account for kinodynamic constraints and robot dynamics: indoor benchmarks can produce unrealistic behaviors such as sliding against obstacles, while outdoor benchmarks often ignore terrain topology such as curbs and steps and assume perfect action execution.
These assumptions can lead deployed policies to collide with obstacles, fall, or choose unsafe terrain.
In contrast, \textsc{NavVerse} uses physics simulation to evaluate realistic action outcomes and safety metrics to quantify safety-aware behavior.

To sum up, our contributions are threefold: i) \textbf{Physics-enabled indoor-to-outdoor benchmark:} We integrate residential, commercial, outdoor, and connected hybrid scenes in continuous robot simulation, with scalable scene/task generation, verified episodes, and executable robot evaluation; ii) \textbf{Rich task spectrum:} We introduce tasks ranging from standard navigation to the novel, long-horizon PlaceNav. These are assessed via task success, path efficiency, and safety-aware execution metrics that expose physical instability and transition failures;
and iii) \textbf{Navigation baseline findings:} We evaluate classical modular, RL, and VLA methods, revealing that representative baselines show some zero-shot transfer to unseen tasks but still struggle with indoor-to-outdoor transitions and safety-aware decision making.

\vspace{-2mm}
\section{Related Work}
\label{sec:related_work}
\vspace{-3mm}

\begin{table*}[ht]
\centering
\setlength{\tabcolsep}{1.8pt}
\caption{Compared to existing benchmarks, NavVerse spans indoor, outdoor, and indoor-to-outdoor environments under continuous robot execution and also offers safety metrics. Comp., Eff., Fail., and Safe denote task completion, efficiency, failure diagnostics, and safety metrics, respectively.}
\vspace{-2mm}
\label{tab:benchmark_comparison}
\resizebox{\textwidth}{!}{%
\begin{tabular}{l l c l c l l l}
\toprule
Benchmark & Scene & Embodiment Dynamics & Action & Scenes & Simulator & Task & Metrics \\
\midrule
Talk The Walk (2018)~\cite{vries_talk_2018} & Outdoor & \xmark & Discrete & 5 & Street View/ParlAI~\cite{vries_talk_2018} & DialogNav & Comp. \\
TouchDown (2019)~\cite{chen_touchdown_2020} & Outdoor & \xmark & Discrete & - & Street View~\cite{chen_touchdown_2020} & VLN & Comp., Eff. \\
R2R~\cite{anderson_vision-and-language_2018}, RxR~\cite{ku_room-across-room_2020} & Indoor & \xmark & Discrete & 90 & MP3D-Sim~\cite{anderson_vision-and-language_2018,mp3d} & VLN & Comp., Eff. \\
RoboTHOR (2020)~\cite{robothor} & Indoor & \xmark & Discrete & 89 & AI2-THOR~\cite{ai2thor} & ObjNav & Comp., Eff. \\
MP3D ObjNav (2020)~\cite{mp3d} & Indoor & \xmark & Discrete & 90 & Habitat~\cite{habitat} & ObjNav & Comp., Eff. \\
HM3D ObjNav (2022)~\cite{habitatchallenge2022} & Indoor & \xmark & Discrete & 120 & Habitat~\cite{habitat} & ObjNav & Comp., Eff. \\
VLN-CE (2020)~\cite{krantz_beyond_2020} & Indoor & \xmark & Continuous & 90 & Habitat~\cite{habitat} & VLN & Comp., Eff. \\
VLN-PE (2025)~\cite{wang_rethinking_2025} & Indoor & \cmark & \makecell[l]{Discrete\\Continuous\\Waypoints} & 100 & Isaac Sim~\cite{NVIDIA_Isaac_Sim} & VLN & \makecell[l]{Comp., Eff.\\Fail.} \\
\midrule
NavVerse (Ours) & \makecell[l]{Indoor\\Outdoor\\Indoor-to-Outdoor} & \cmark & \makecell[l]{Discrete\\Continuous\\Waypoints} & 200 & Isaac Sim~\cite{NVIDIA_Isaac_Sim} & \makecell[l]{ObjNav\\VLN\\PlaceNav} & \makecell[l]{Comp., Eff.\\Fail., Safe.} \\
\bottomrule
\end{tabular}%
}
\vspace{-3mm}
\end{table*}

\noindent\textbf{Simulation platform.} Existing simulators generally prioritize either visual or physics fidelity. Visual-fidelity platforms: Platforms like Habitat-Sim~\cite{habitat, habitat2, habitat3} (utilizing scanned environments like Matterport3D~\cite{mp3d} and HM3D~\cite{hm3d}) offer high-throughput egocentric rendering for vision-based navigation, while AI2-THOR~\cite{ai2thor} supports interactive indoor scenes. However, benchmarks on these platforms often simplify physical execution using discrete actions or lightweight dynamics. Physics-fidelity platforms: Frameworks like MuJoCo~\cite{mujoco} and Gazebo~\cite{gazebo} excel at continuous control and contact dynamics but lack visually rich, large-scale navigation environments. To bridge this gap, Isaac Sim~\cite{NVIDIA_Isaac_Sim} serves as a powerful backend by combining GPU-accelerated visual rendering with articulated physics and continuous control. Built upon Isaac Sim, our NavVerse delivers both visual and physics fidelity, offering a comprehensive embodied navigation platform that unifies photorealistic scenes with executable robot motion, terrain annotations, and diagnostic metrics.

\noindent\textbf{Indoor and outdoor navigation benchmarks.} 
Most existing navigation benchmarks study indoor and outdoor navigation as separate settings, so the transition between the two remains less explored. 
Indoor benchmarks have mainly studied Object Navigation (ObjNav)~\cite{Gibson, mp3d, hm3d,procthor,habitat, habitat2, habitat3, robothor}, and Vision-and-Language Navigation (VLN)~\cite{anderson_vision-and-language_2018, ku_room-across-room_2020, krantz_beyond_2020, wang_rethinking_2025, thomason_vision-and-dialog_2019, qiao_navbench_2025, alfred,zhu_soon_2021, qi_reverie_2020, wang_navrag_2025}. 
These benchmarks have provided strong testbeds for indoor semantic understanding and instruction following, but they are typically centered on indoor scenes rather than connected transitions to outdoor space.
Outdoor benchmarks often rely on street-view imagery and panoramic observations~\cite{vries_talk_2018, chen_touchdown_2020}, or pre-recorded trajectories~\cite{schumann_generating_2021, li_vln-video_2024, liang_gnd_2025}. Virtual Community~\cite{zhou_virtual_nodate} moves toward simulated outdoor environments using Google 3D Tiles and street-view imagery. 
These benchmarks extend navigation evaluation to larger-scale environments, but many of them are built around panoramic observations and non-interactive navigation interfaces, which makes it difficult to study continuous robot execution, safety, and embodiment-dependent traversability.
To bridge these gaps, NavVerse connects diverse interactive scenes (indoor, outdoor and indoor-to-outdoor) with continuous robot-execution protocol, offering a comprehensive embodied navigation simulation platform, see Table~\ref{tab:benchmark_comparison}. It pioneers a new research direction centered on the practical challenges of indoor-to-outdoor transitions.

\vspace{-2mm}
\section{NavVerse: Indoor-to-Outdoor Navigation Benchmark}
\vspace{-2mm}

\begin{figure*}
    \centering
    \includegraphics[width=0.8\linewidth]{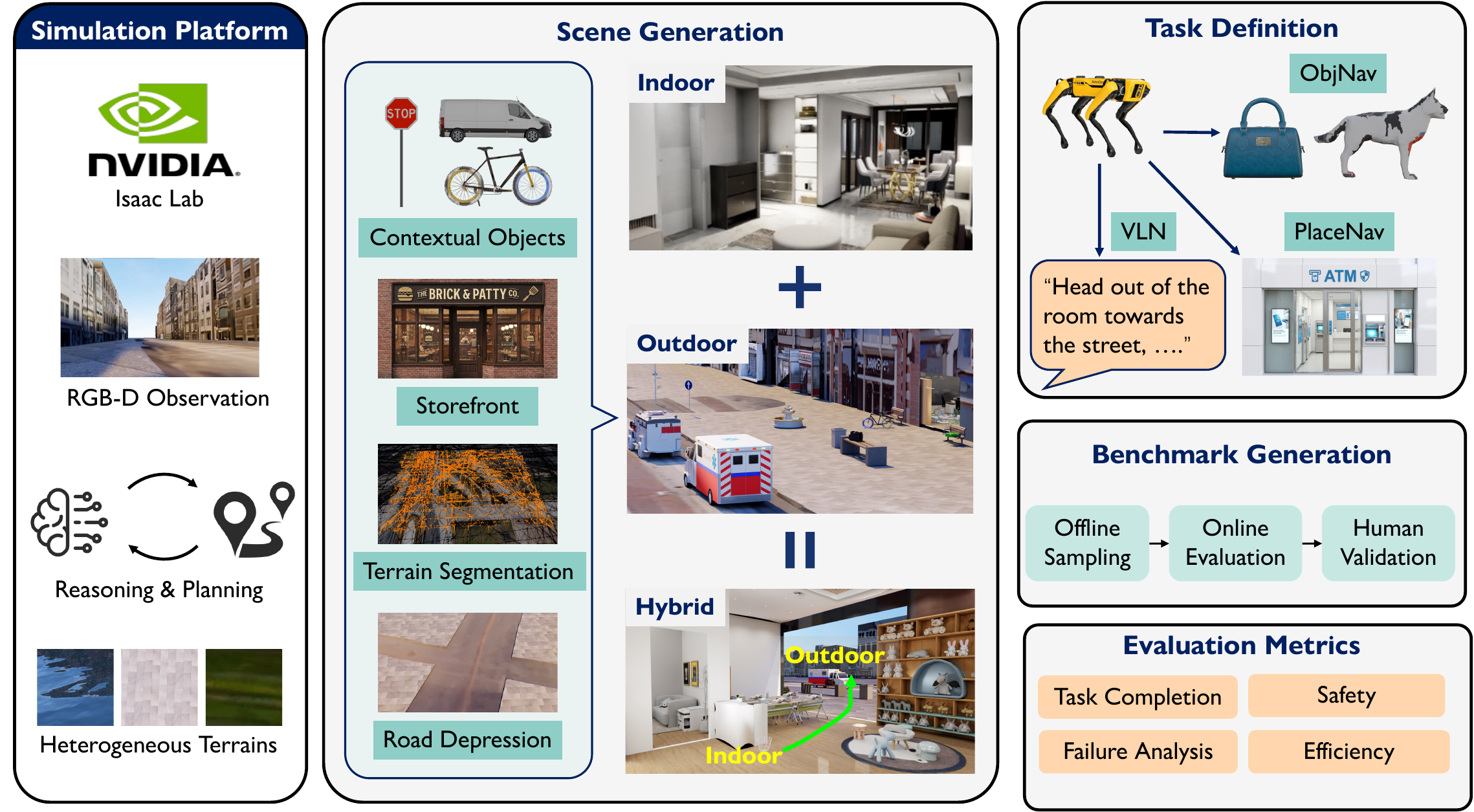}
    \caption{Overview of the \textsc{NavVerse} benchmark construction pipeline. Isaac Sim provides the execution engine, while \textsc{NavVerse} contributes connected indoor, outdoor, and indoor-to-outdoor scene construction, task generation, oracle validation, and diagnostic evaluation.}
    \label{fig:benchmark}
    \vspace{-8mm}
\end{figure*}

In this section, we introduce \textsc{NavVerse}, a comprehensive embodied navigation benchmark, spanning indoor, outdoor and indoor-to-outdoor scenarios. The benchmark construction overview is shown in Fig.~\ref{fig:benchmark}, including scene generation, task definition and evaluation. Note that NavVerse is built upon Isaac Sim, offering both photorealistic scenes and physics-aware robot execution.

\vspace{-2mm}
\subsection{Scene Generation}
\label{sec:scene_generation}
\vspace{-2mm}
\textsc{NavVerse} covers indoor, outdoor, and indoor-to-outdoor scenes. To ensure a robust evaluation of generalization capabilities, the training, and evaluation splits feature mutually exclusive scenes. We defer implementation details for asset generation, mesh processing, object placement, storefront construction, terrain modeling, and indoor-to-outdoor alignment to Appendix~\ref{sec:app_datasets_scenes}.

\noindent\textbf{Indoor Scenes.}
Our indoor scenes are built upon\textit{GRScenes} meshes from  GRUtopia~\cite{wang2024grutopia}. To promote scene diversity, we curate 100 scenes covering apartments and commercial spaces such as groceries, hospitals, and offices.
These indoor scenes feature diverse object categories and cluttered layouts, demanding both semantic reasoning and physics-aware collision avoidance capabilities.

\noindent\textbf{Outdoor Scenes.}
Our outdoor scenes are built on 50 meshes from \textit{Virtual Community}~\cite{zhou_virtual_nodate}. We enhance these meshes across several key dimensions, transforming sparse urban geometry into semantically rich, physics-backed, and interactive environments.

\noindent$\bullet$ \textbf{Real-to-sim urban layouts and physics-aware terrains.} The scenes preserve real-world spatial layouts and urban scale through Google 3D Tiles geometry and map-aligned urban structure. Furthermore, to enable kinodynamic constraints and surface-dependent traversability, we segment terrain into map-derived material classes and refine the geometry of roads, sidewalks, and open areas with realistic elevation changes. This allows our benchmark to evaluate true trajectory feasibility rather than simple semantic goal recognition or topological routing.

\noindent$\bullet$ \textbf{Rich object and photorealistic storefronts population.} We populate the layout lanes and sidewalks with semantically reasonable objects sampled from 121 Objaverse \cite{deitke2023objaverse} object categories, supporting up to 200 cars and 1,000 sidewalk objects per scene. To support PlaceNav and language-guided tasks, we also generate and place 101 diverse storefront assets (restaurants, cafes, banks) at map-aligned building frontages.

\noindent\textbf{Indoor-to-Outdoor Scenes.}
To evaluate connected cross-context navigation, we integrate indoor scenes into buildings in the outdoor world, resulting in 50 indoor-to-outdoor scenes.

\noindent$\bullet$ \textbf{Door-to-facade scene assembly.} We construct indoor-to-outdoor scenes with two key steps: i) identify a road-facing building and create a facade opening as the exit; and ii) place a compatible indoor layout behind the building. To align the indoor entrance with the opening, we need to adjust the entrance height, and remove blocking door or wall meshes to connect the spaces.

\noindent$\bullet$ \textbf{Consistent physical execution.} The connected indoor and outdoor components share the same robot-execution protocol, physics setup, sensing, and control stack across the full episode.

These indoor-to-outdoor navigation scenes task agents with traversing corridors, identifying exits, and crossing into open streets toward outdoor goals. This introduces unprecedented challenges for embodied navigation, requiring rapid search adaptation amid drastic shifts in lighting, appearance, environmental structure, and spatial scale.

\vspace{-2mm}
\subsection{Task Definition} 
\vspace{-2mm}

\textsc{NavVerse} provides comprehensive evaluation across three task types under three scene types.

\medskip\noindent\textbf{Object Navigation.}
Object navigation (ObjNav) requires the agent to locate and approach specific object categories within the environment. 
While indoor scenes focus on fine-grained object recognition, outdoor scenes challenge the agent to identify landmarks over much larger distances.

\medskip\noindent\textbf{Place Navigation.}
Place navigation (PlaceNav) requires agents to reach functional destinations, such as restaurants, cafes, or banks, rather than localized visual objects. Embedded within the urban layout, PlaceNav challenges agents to reason over road topology, building frontages, storefront semantics, and long-range visibility cues to locate the target destination.

\medskip\noindent\textbf{Vision-and-Language Navigation.}
In Vision-and-Language Navigation (VLN), the agent is tasked to follow natural language instructions to reach a goal. 
This evaluates the agent's ability to ground linguistic cues in visual observations and execute complex, multi-step trajectories. 

\vspace{-2mm}
\subsection{Benchmark and Evaluation}
\vspace{-2mm}

\noindent\textbf{Episode generation.} \textsc{NavVerse} generates episodes with three steps: i) \textit{offline sampling} of start-to-goal paths on the navigation mesh (NavMesh); ii) \textit{online filtering} via simulated robot execution to discard infeasible paths and record successful sensor observations and poses; and iii) \textit{human validation} utilizing VLMs for VLN instruction generation and a web interface for manual quality control. This process guarantees high-quality, executable episodes, culminating in a benchmark of 4{,}027 ObjNav episodes, 2{,}973 PlaceNav episodes, and 3{,}000 VLN episodes. Detailed episode numbers by scene type and task are provided in Table~\ref{tab:episode_counts}.

\noindent\textbf{Evaluation setup.}
We evaluate all models on the evaluation split, reporting both full task-level summaries and domain-specific diagnostic subsets for ObjNav and PlaceNav. Experiments utilize a simulated Boston Dynamics Spot robot with an RGB-D sensor. To preserve each baseline's native structure, model outputs are converted to the \textsc{NavVerse} interface and executed using a shared waypoint follower managed by a PID controller. Detailed observation, action, control, and episode specifications are provided in Appendix~\ref{sec:app_obs}--\ref{sec:app_episodes}.

\noindent\textbf{Evaluation metrics.}
We report metrics along three axes: task completion, exploration efficiency, and safety. 
Task completion includes Success Rate (SR) and Success weighted by Path Length (SPL). 
Exploration efficiency includes Coverage Efficiency (CE), which measures the amount of newly covered space per unit travel distance and is especially important for large outdoor and indoor-to-outdoor scenes. 
Safety metrics include Collision Rate (CR), Average Distance to Obstacles (ADO), and Navigable Surface Ratio (NSR): CR quantifies collision with obstacles, ADO measures clearance from nearby obstacles, and NSR measures whether the robot remains on valid or kinodynamically preferred navigable surfaces. 
Success uses a geometric goal tolerance of $1.0+r_{\mathrm{robot}}$ meters, which is $1.6$ m for the Spot robot used in our experiments. 
Episodes terminate on success, timeout, fall, or stop at wrong goal.
Formal metric definitions are provided in Appendix~\ref{sec:app_metrics}.

\vspace{-2mm}
\section{Experiments}
\label{sec:experiment}
\vspace{-2mm}

\subsection{Baselines}
\vspace{-1mm}

We evaluate four representative baselines.
SGImagineNav~\cite{hu_imaginative_2025} represents modular map-based semantic navigation; we connect its planned waypoints to the \textsc{NavVerse} controller.
PoliFormer~\cite{zeng_poliformer_2024} is an indoor RL policy and tests transfer from indoor training to heterogeneous indoor/outdoor execution.
UniNaVid~\cite{uninavid} is a VLA navigation model evaluated across object-, place-, and language-conditioned tasks.
LongNav-R1~\cite{hu_longnav-r1_2026} is an end-to-end VLA navigation policy optimized via RL within HM3D ObjNav~\cite{habitatchallenge2022}.
All methods are evaluated under the same physics-enabled execution protocol.
SGImagineNav directly uses the \textsc{NavVerse} waypoint interface, while PoliFormer, UniNaVid, and LongNav-R1 output discrete actions that are converted into single-step waypoints and executed by our shared waypoint follower.
For VLN, we report UniNaVid as the only baseline among the evaluated methods that natively accepts route-style language instructions.

\vspace{-1mm}
\subsection{Multiple Navigation Task Evaluation}
\label{sec:multi_task_eval}
\vspace{-1mm}

To provide a comprehensive evaluation of existing methods, we report success rate and path efficiency in Table~\ref{tab:result_task_scene}, and fine-grained safety and efficiency metrics in Table~\ref{tab:allscene_metrics_main}.

\begin{table*}[t]
\centering
\caption{\textbf{Navigation success rate and path efficiency across indoor, outdoor and indoor-to-outdoor scenes.}
PlaceNav has no indoor-only split because place-level goals are defined over outdoor POIs and indoor-to-outdoor episodes.
Bold marks the best result.}
\label{tab:result_task_scene}
\footnotesize
\setlength{\tabcolsep}{3pt}
\renewcommand{\arraystretch}{0.90}
\begin{adjustbox}{max width=0.9\textwidth}
\begin{tabular}{lllcccccccc}
\toprule
\multirow{2}{*}{Task} & \multirow{2}{*}{Method} & \multirow{2}{*}{Category}
& \multicolumn{2}{c}{\textbf{All Scenes}}
& \multicolumn{2}{c}{\textbf{Indoor}}
& \multicolumn{2}{c}{\textbf{Outdoor}}
& \multicolumn{2}{c}{\textbf{Indoor-to-Outdoor}} \\
\cmidrule(lr){4-5}\cmidrule(lr){6-7}\cmidrule(lr){8-9}\cmidrule(lr){10-11}
&&& SR$\uparrow$ & SPL$\uparrow$ & SR$\uparrow$ & SPL$\uparrow$ & SR$\uparrow$ & SPL$\uparrow$ & SR$\uparrow$ & SPL$\uparrow$ \\
\midrule
\multirow{4}{*}{ObjNav}
& SGImagineNav~\cite{hu_imaginative_2025} & Modular & 6.42  & 0.61 & 10.00 & 0.73 & 2.44 & 0.21 & 4.21 & 0.74 \\
& PoliFormer~\cite{zeng_poliformer_2024}  & RL      & 5.81  & 1.00 & 12.67 & 2.17 & 0.00 & 0.00 & 0.00 & 0.00 \\
& UniNaVid~\cite{uninavid}                & VLA     & \textbf{11.62} & \textbf{2.58} & \textbf{17.33} & \textbf{3.82} & \textbf{4.88} & \textbf{1.30} & \textbf{8.42} & \textbf{1.71} \\
& LongNav-R1~\cite{hu_longnav-r1_2026}    & VLA-RL  & 4.59  & 0.62 & 6.67  & 0.89 & 2.44 & 0.28 & 3.16 & 0.47 \\
\midrule
\multirow{4}{*}{PlaceNav}
& SGImagineNav~\cite{hu_imaginative_2025} & Modular & 4.88  & 0.64 & -- & -- & 5.88 & 0.90 & \textbf{3.64} & 0.32 \\
& PoliFormer~\cite{zeng_poliformer_2024}  & RL      & 1.63  & 0.03 & -- & -- & 2.94 & 0.05 & 0.00 & 0.00 \\
& UniNaVid~\cite{uninavid}                & VLA     & \textbf{11.38} & \textbf{3.30} & -- & -- & \textbf{17.65} & \textbf{4.98} & \textbf{3.64} & \textbf{1.22} \\
& LongNav-R1~\cite{hu_longnav-r1_2026}    & VLA-RL  & 6.50  & 1.23 & -- & -- & 8.82 & 1.92 & \textbf{3.64} & 0.38 \\
\midrule
VLN
& UniNaVid~\cite{uninavid}                & VLA     & 10.67 & 3.21 & 10.67 & 2.04 & 14.00 & 5.31 & 7.33 & 2.29 \\
\bottomrule
\end{tabular}
\end{adjustbox}
\vspace{-2mm}
\end{table*}

\begin{table*}[t]
\centering
\setlength{\tabcolsep}{3pt}
\begin{minipage}[t]{0.40\textwidth}
\vspace{0pt} 
\caption{\textbf{Safety and efficiency diagnostics.} Lower is better for Collision Rate; higher is better for others.}
\label{tab:allscene_metrics_main}
\centering
\resizebox{\textwidth}{!}{%
\begin{tabular}{llcccc}
\toprule
\multirow{2}{*}{Task} & \multirow{2}{*}{Method} & \multicolumn{3}{c}{\textbf{Safety}} & \textbf{Efficiency} \\ \cmidrule(lr){3-5} \cmidrule(lr){6-6}
 &  & CR$\downarrow$ & ADO$\uparrow$ & NSR$\uparrow$ & CE$\uparrow$ \\ \midrule
\multirow{4}{*}{ObjNav} & PoliFormer & 0.36 & 0.72 & 0.10 & 0.37 \\
 & UniNaVid & 0.31 & 0.81 & 0.12 & \textbf{0.52} \\
 & SGImagineNav & \textbf{0.21} & 0.98 & \textbf{0.16} & 0.48 \\
 & LongNav-R1 & 0.34 & \textbf{1.17} & 0.13 & 0.50 \\ \midrule
\multirow{4}{*}{PlaceNav} & PoliFormer [8] & 0.24 & 1.56 & 0.09 & 0.38 \\
 & UniNaVid & 0.16 & 1.98 & 0.12 & 0.62 \\
 & SGImagineNav & \textbf{0.05} & 1.91 & \textbf{0.13} & 0.49 \\
 & LongNav-R1 & 0.20 & \textbf{2.21} & 0.11 & \textbf{0.65} \\ \midrule
VLN & UniNaVid & 0.24 & 1.28 & 0.14 & 0.45 \\ \bottomrule
\end{tabular}%
}
\end{minipage}
\hfill 
\begin{minipage}[t]{0.57\textwidth}
\vspace{0pt} 
\caption{\textbf{Stage-wise transition analysis on indoor-to-outdoor episodes.} Ind-Only: agent remained indoors; Reach-Out: agent crossed outdoors; Wrong-G: navigated to an incorrect goal.}
\label{tab:stagewise_transition_success}
\centering
\resizebox{\textwidth}{!}{%
\begin{tabular}{llccccccc}
\toprule
\multirow{2}{*}{Task} & \multirow{2}{*}{Method} & \multicolumn{2}{c}{Transition Stage} & \multicolumn{2}{c}{Efficiency (CE$\uparrow$)} & \multicolumn{3}{c}{Post-Exit Failure Cases} \\ \cmidrule(lr){3-4} \cmidrule(lr){5-6} \cmidrule(lr){7-9}
 &  & Ind-Only $\downarrow$ & Reach-Out$\uparrow$ & Pre-Exit & Post-Exit & Wrong-G & Fall & Time Out \\ \midrule
\multirow{4}{*}{ObjNav} & PoliFormer & 46.32 & 53.68 & 0.89 & 0.66 & 5.9 & 17.6 & 76.5 \\
 & UniNaVid & 48.42 & 51.58 & \textbf{1.07} & 0.84 & 54.8 & 4.8 & 40.5 \\
 & SGImagineNav & \textbf{25.26} & \textbf{74.74} & 0.74 & 0.49 & 26.1 & 7.2 & 66.7 \\
 & LongNav-R1 & 47.37 & 52.63 & 1.01 & \textbf{0.86} & 74.5 & 8.5 & 17.0 \\ \midrule
\multirow{4}{*}{PlaceNav} & PoliFormer & 32.73 & 67.27 & 0.79 & 0.62 & 5.4 & 24.3 & 70.3 \\
 & UniNaVid & 38.18 & 61.82 & \textbf{1.26} & \textbf{0.81} & 48.5 & 6.1 & 45.5 \\
 & SGImagineNav & \textbf{30.91} & \textbf{69.09} & 0.61 & 0.53 & 8.3 & 22.2 & 69.4 \\
 & LongNav-R1 & \textbf{30.91} & \textbf{69.09} & 0.88 & 0.79 & 72.2 & 13.9 & 13.9 \\ \midrule
VLN & UniNaVid & 37.33 & 62.67 & 1.14 & 0.80 & 73.8 & 1.2 & 25.0 \\ \bottomrule
\end{tabular}%
}
\end{minipage}
\vspace{-4mm}
\end{table*}

\medskip\noindent\textbf{UniNaVid (VLA) achieves the best zero-shot task-completion performance.}
Table~\ref{tab:result_task_scene} shows that UniNaVid achieves the strongest task-completion performance among comparable zero-shot baselines on both ObjNav and PlaceNav.
However, the low absolute SR across all tasks, including 10.67\% SR on UniNaVid VLN, shows that physically grounded navigation remains far from solved.

\medskip\noindent\textbf{Existing methods demonstrate divergent success, efficiency, and safety behaviors.} 
Tables~\ref{tab:result_task_scene} and \ref{tab:allscene_metrics_main} reveal that existing methods excel only in orthogonal dimensions: the VLA model \textsc{UniNaVid} leads in task success and search efficiency (SR, SPL, CE), \textsc{LongNav-R1} ensures safety via obstacle clearance (high ADO), and \textsc{SGImagineNav} optimizes local interaction safety (low CR, high NSR). This divergence underscores that current baselines remain insufficient for holistic deployment. Crucially, NavVerse isolates these architectural blind spots, motivating the development of next-generation models capable of simultaneously balancing success, safety, and efficiency, all of which are non-negotiable requirements for real-world navigation.

\medskip\noindent\textbf{PlaceNav presents more pronounced indoor-to-outdoor transition challenges than VLN and ObjNav.} As shown in Table~\ref{tab:result_task_scene}, the performance of the leading model (\textsc{UniNaVid}) degrades sharply when transitioning to indoor-to-outdoor episodes. Success rates drop by 8.91\% for ObjNav (17.33\% to 8.42\%) and by 6.67\% for VLN (14.00\% to 7.33\%). The degradation is most severe in PlaceNav, where performance plummets by an absolute 14.01\%, dropping from 17.65\% in pure outdoor scenes to a mere 3.64\% on indoor-to-outdoor episodes. The reasons are twofold: i) VLN primarily requires instruction-following capabilities that translate effectively across domains, resulting in the most stable performance across environments; ii) conversely, PlaceNav necessitates topology-aware point-of-interest searching, which diverges sharply between indoor and outdoor layouts, imposing much steeper adaptation bottlenecks than ObjNav.

\vspace{-1mm}
\subsection{Indoor-to-Outdoor Transition Gap}
\label{sec:transition_gap}
\vspace{-1mm}

To understand the challenges of indoor-to-outdoor transitions, we conduct a detailed failure mode analysis in Tables~\ref{tab:stagewise_transition_success} and \ref{tab:termination_reasons}. This evaluation pinpoints failures across the complete navigation sequence, which requires an agent to start indoors, reason over structural transitions, locate an exit, cross the boundary, and execute an outdoor goal-conditioned search.

\begin{table*}[t]
\centering
\caption{\textbf{Failure case analysis.}
Entries are percentages over failed episodes. \#Fail denotes the number of unsuccessful episodes.}
\vspace{-2mm}
\label{tab:termination_reasons}
\scriptsize
\setlength{\tabcolsep}{2.0pt}
\renewcommand{\arraystretch}{0.88}
\begin{adjustbox}{max width=\textwidth}
\begin{tabular}{ll cccc cccc cccc}
\toprule
\multirow{2}{*}{Task} & \multirow{2}{*}{Method}
& \multicolumn{4}{c}{Indoor}
& \multicolumn{4}{c}{Outdoor}
& \multicolumn{4}{c}{Indoor-to-Outdoor} \\
\cmidrule(lr){3-6}\cmidrule(lr){7-10}\cmidrule(lr){11-14}
&& \#Fail & Wrong Goal & Fall & Timeout
& \#Fail & Wrong Goal & Fall & Timeout
& \#Fail & Wrong Goal & Fall & Timeout \\
\midrule
\multirow{4}{*}{ObjNav}
& PoliFormer~\cite{zeng_poliformer_2024}
& 131 & 9.2  & 31.3 & 59.5
& 82  & 2.4  & 2.4  & 95.1
& 95  & 6.3  & 23.2 & 70.5 \\
& UniNaVid~\cite{uninavid}
& 124 & 52.4 & 16.9 & 30.6
& 78  & 55.1 & 9.0  & 35.9
& 87  & 35.6 & 20.7 & 43.7 \\
& SGImagineNav~\cite{hu_imaginative_2025}
& 135 & 33.3 & 43.0 & 23.7
& 80  & 26.2 & 7.5  & 66.2
& 91  & 29.7 & 15.4 & 54.9 \\
& LongNav-R1~\cite{hu_longnav-r1_2026}
& 140 & 56.4 & 22.9 & 20.7
& 80  & 95.0 & 2.5  & 2.5
& 92  & 48.9 & 25.0 & 26.1 \\
\midrule
\multirow{4}{*}{PlaceNav}
& PoliFormer~\cite{zeng_poliformer_2024}
& -- & -- & -- & --
& 66 & 1.5  & 13.6 & 84.8
& 55 & 5.5  & 27.3 & 67.3 \\
& UniNaVid~\cite{uninavid}
& -- & -- & -- & --
& 56 & 33.9 & 7.1  & 58.9
& 53 & 39.6 & 5.7  & 54.7 \\
& SGImagineNav~\cite{hu_imaginative_2025}
& -- & -- & -- & --
& 64 & 1.6  & 18.8 & 79.7
& 53 & 7.5  & 24.5 & 67.9 \\
& LongNav-R1~\cite{hu_longnav-r1_2026}
& -- & -- & -- & --
& 62 & 87.1 & 9.7  & 3.2
& 53 & 69.8 & 15.1 & 15.1 \\
\midrule
VLN
& UniNaVid~\cite{uninavid}
& 134 & 26.9 & 23.9 & 49.3
& 129 & 82.9 & 5.4  & 11.6
& 139 & 59.7 & 7.2  & 33.1 \\
\bottomrule
\end{tabular}
\end{adjustbox}
\vspace{-4mm}
\end{table*}

\medskip\noindent\textbf{Indoor and outdoor scenes stress distinct failure modes, shifting from local execution challenges indoors to exploration and goal-grounding limits outdoors.} Table~\ref{tab:termination_reasons} indicates that indoor ObjNav failures are dominated by local execution constraints in confined spaces: falls cause 31.3\% of PoliFormer and 43.0\% of SGImagineNav failures. Outdoors, the bottleneck shifts to exploration and grounding. PoliFormer times out in 95.1\% of failed outdoor ObjNav and 84.8\% of PlaceNav episodes, while LongNav-R1 mostly fails due to incorrect goals (95.0\%/87.1\% on ObjNav/PlaceNav). And 82.9\% of UniNaVid outdoor VLN failures are wrong-goal issues. Post-exit metrics in Table~\ref{tab:stagewise_transition_success} mirror these trends, confirming that these vulnerabilities persist after transitioning outdoors.

\medskip\noindent\textbf{Indoor-to-outdoor episodes introduce an exit-finding bottleneck.}
Table~\ref{tab:stagewise_transition_success} shows that many indoor-to-outdoor episodes fail before the agent reaches the outdoor side.
On ObjNav, UniNaVid and LongNav-R1 are Not Reached Outdoor in 48.42\% and 47.37\% of episodes, respectively.
Even SGImagineNav, which achieves the highest reached-outdoor rate, still fails to reach outdoor space in 25.26\% of ObjNav episodes and 30.91\% of PlaceNav episodes.
For VLN, UniNaVid is also Not Reached Outdoor in 37.33\% of episodes.
This failure mode is invisible in pure indoor or pure outdoor evaluation, showing the importance of indoor-to-outdoor scenes.

\medskip\noindent\textbf{Reaching outdoor space causes a systematic efficiency drop.}
Table~\ref{tab:stagewise_transition_success} further shows that all methods have lower coverage efficiency after the exit step than before it.
For example, UniNaVid drops from 1.07 to 0.84 on ObjNav, from 1.26 to 0.81 on PlaceNav, and from 1.14 to 0.80 on VLN; the same pre-to-post exit CE drop appears for the other ObjNav and PlaceNav methods.
We observe that after exiting the indoor area, agents often exhibit inefficient behaviors such as persistent in-place rotation and back-and-forth motion. This suggests limited adaptation to changes in scene scale and topology: although outdoor regions provide more open space to explore, agents fail to adjust their exploration strategy and cover the space less efficiently after reaching outdoor space.

\medskip\noindent\textbf{Post-exit failures are method-specific.}
Among outdoor-reached failures, different methods fail for different reasons.
PoliFormer and SGImagineNav are mainly limited by timeouts after reaching outdoor space.
LongNav-R1 instead mostly fails through wrong-goal termination, with wrong-goal failures accounting for 74.5\% on ObjNav and 72.2\% on PlaceNav, suggesting that it often calls \texttt{STOP} without sufficient evidence that it has reached the correct goal.
UniNaVid shows a mixed pattern: wrong-goal failures are prominent, especially on VLN, but timeout remains substantial on ObjNav and PlaceNav.
These results indicate that reaching outdoor space is not sufficient; agents also need to adapt their exploration strategy and maintain reliable goal grounding after the exit step.

\vspace{-1mm}
\subsection{Diagnostic Analyses}
\label{sec:diagnostic_analyses}
\vspace{-1mm}

\begin{table*}[t]
\centering
\begin{minipage}[t]{0.53\textwidth}
\vspace{0pt} 
\caption{\textbf{Performance by episode difficulty.} Easy, Medium, and Hard difficulty levels correspond to increasing path lengths.}
\vspace{-1mm}
\label{tab:result_task_difficulty}
\centering
\resizebox{\textwidth}{!}{%
\begin{tabular}{llcccccc}
\toprule
\multirow{2}{*}{Task} & \multirow{2}{*}{Method} & \multicolumn{2}{c}{\textbf{Easy}} & \multicolumn{2}{c}{\textbf{Medium}} & \multicolumn{2}{c}{\textbf{Hard}} \\ \cmidrule(lr){3-4} \cmidrule(lr){5-6} \cmidrule(lr){7-8} 
 &  &  SR$\uparrow$ & SPL$\uparrow$ & SR$\uparrow$ & SPL$\uparrow$ & SR$\uparrow$ & SPL$\uparrow$ \\ \midrule
\multirow{4}{*}{ObjNav} & PoliFormer & 15.60 & 2.76 & 1.83 & 0.23 & 0.00 & 0.00 \\
 & UniNaVid  & \textbf{21.10} & \textbf{4.98} & \textbf{11.93} & \textbf{2.53} & \textbf{1.83} & 0.22 \\
 & SGImagineNav & 11.93 & 0.84 & 5.50 & 0.48 & \textbf{1.83} & \textbf{0.50} \\
 & LongNav-R1 & 8.26 & 1.09 & 4.59 & 0.63 & 0.92 & 0.13 \\ \midrule
\multirow{4}{*}{PlaceNav} & PoliFormer  & 4.88 & 0.08 & 0.00 & 0.00 & 0.00 & 0.00 \\
 & UniNaVid & \textbf{14.63} & \textbf{3.23} & \textbf{14.63} & \textbf{5.57} & \textbf{4.88} & 1.10 \\
 & SGImagineNav & 4.88 & 0.22 & 4.88 & 0.97 & \textbf{4.88} & 0.74 \\
 & LongNav-R1 & 9.76 & 0.96 & 7.32 & 1.33 & 2.44 & \textbf{1.39} \\ \midrule
VLN & UniNaVid & 13.33 & 2.74 & 14.00 & 4.49 & 4.67 & 2.41 \\ \bottomrule
\end{tabular}%
}
\end{minipage}
\hfill
\begin{minipage}[t]{0.44\textwidth}
\caption{\textbf{Impact of kinodynamics under oracle-trajectory execution.} }
\vspace{-1mm}
\label{tab:robot_performance_main}
\centering
\resizebox{\textwidth}{!}{%
\begin{tabular}{lcccc}
\toprule
\multirow{2}{*}{Friction ($\mu$)} & \multicolumn{2}{c}{Legged (Spot)} & \multicolumn{2}{c}{Wheeled (Turtlebot)} \\ \cmidrule(lr){2-3} \cmidrule(lr){4-5} 
 & SR$\uparrow$ & Vel. (m/s)$\uparrow$ & SR$\uparrow$ & Vel. (m/s)$\uparrow$ \\ \midrule
$\mu = 0.5$ & 100.00 & 1.46 & 30.75 & 0.36 \\
$\mu = 1.0$ & 100.00 & 1.49 & 47.50 & 0.58 \\ \bottomrule
\end{tabular}%
}
\vspace{1mm}

\caption{\textbf{Impact of instruction (PlaceNav).}}
\vspace{-1mm}
\label{tab:instruction_type_placenav}
\centering
\resizebox{\textwidth}{!}{%
\begin{tabular}{lcccccc}
\toprule
\multirow{2}{*}{Instruction Type} & \multicolumn{2}{c}{\textbf{All}} & \multicolumn{2}{c}{\textbf{Outdoor}} & \multicolumn{2}{c}{\textbf{Indoor-to-Outdoor}} \\ \cmidrule(lr){2-3} \cmidrule(lr){4-5} \cmidrule(lr){6-7} 
 & SR$\uparrow$ & SPL$\uparrow$ & SR$\uparrow$ & SPL$\uparrow$ & SR$\uparrow$ & SPL$\uparrow$ \\ \midrule
Store Category & 11.38 & 3.30 & 17.65 & 4.98 & \textbf{3.64} & \textbf{1.22} \\
VLN Instruction & \textbf{12.20} & \textbf{4.84} & \textbf{19.12} & \textbf{7.89} & \textbf{3.64} & 1.06 \\
Intention Driven & 4.07 & 0.66 & 7.35 & 1.20 & 0.00 & 0.00 \\ \bottomrule
\end{tabular}%
}
\end{minipage}
\vspace{-4mm}
\end{table*}

\medskip\noindent\textbf{Longer routes expose compounding errors.}
Table~\ref{tab:result_task_difficulty} groups episodes into task-specific path-length tertiles, yielding balanced Easy, Medium, and Hard splits within each task.
Across ObjNav, PlaceNav, and VLN, the Hard split consistently reduces success for the strongest baselines compared with the easier splits.
This suggests that long-horizon navigation in \textsc{NavVerse} is limited not only by goal recognition, but also by compounding errors in exploration, memory, terrain selection, instruction grounding, and closed-loop control.
Difficulty thresholds are computed separately within each task using the $1/3$ and $2/3$ quantiles of path length. For ObjNav, the thresholds are 8.07~m and 28.28~m; for PlaceNav, 24.66~m and 39.21~m; and for VLN, 11.61~m and 32.61~m.

\medskip\noindent\textbf{Embodiment dynamics matter even for oracle paths.}
Table~\ref{tab:robot_performance_main} provides a controlled oracle-trajectory analysis over embodiment and friction.
The legged robot completes the tested oracle trajectories under both friction settings, while the wheeled robot has substantially lower SR and velocity.
This diagnostic supports the need for physics-enabled evaluation: a navigation benchmark that ignores embodiment dynamics can overestimate kinodynamic executability under realistic robot embodiments.

\medskip\noindent\textbf{Instruction type affects efficiency and semantic grounding.}
Table~\ref{tab:instruction_type_placenav} compares UniNaVid on the same PlaceNav episodes under three instruction formats.
Route-style VLN instructions modestly improve efficiency over Store Category prompts, increasing overall SPL from 3.30\% to 4.84\%, mainly through outdoor episodes.
However, they do not improve indoor-to-outdoor SR, which remains 3.64\%, suggesting that route descriptions alone do not solve exit finding or post-transition adaptation.
Intention Driven instructions introduce an additional intent-to-place grounding step, reducing overall SR to 4.07\% and indoor-to-outdoor SR to 0.00\%.
\vspace{-2mm}
\section{Conclusion}
\vspace{-2mm}
We introduced \textit{NavVerse}, a physics-enabled navigation benchmark that unifies indoor, outdoor, and indoor-to-outdoor evaluation under a consistent protocol. NavVerse supports three complementary tasks, Object Navigation, Place Navigation, and Vision-and-Language Navigation, to stress goal conditioned reaching, long horizon semantic search, and instruction grounding and following. Experiments across RL, end-to-end VLA, and modular baselines reveal a clear indoor-to-outdoor transition gap, motivating methods that are robust to perception shifts, terrain variability, spatial re-anchoring, and long-horizon semantics in cross-context navigation settings.

\noindent\textbf{Limitations and future work.} NavVerse has several limitations that suggest concrete next steps. First, most environments are currently single floor meshes. Future direction should expand to multi floor layouts and overpass or skybridge structures to better reflect realistic urban navigation and vertical transitions.  Second, the benchmark currently contains limited dynamics and interaction. We aim to introduce more dynamic agents such as pedestrians and vehicles, and more interactive objects such as doors and movable obstacles, to enable evaluation of socially aware and interaction aware navigation. Finally, outdoor-to-indoor transition is not explored in this work, we leave this setting to future extensions.

\clearpage
\appendix

\section{Benchmark Interface and Evaluation Protocol}
\label{sec:app_benchmark}

\subsection{Observations}
\label{sec:app_obs}
At each decision step, the agent receives onboard sensing and proprioception:
\begin{itemize}
  \item \textbf{RGB-D.} We provide a forward-facing RGB image and a depth image rendered at $640\times480$ and $10$ FPS.
  The depth range is $[0.1, 1000]$ meters. We do not add an additional depth noise model.
  \item \textbf{Robot state and geometry.} We provide the ground-truth robot base pose, the camera-to-body transforms, camera intrinsics, and robot size. We do not provide IMU measurements.
\end{itemize}
RGB, depth, and pose are published together every time an image is rendered.
Most experiments in this paper use the Boston Dynamics Spot robot as the primary embodiment.
We also instantiate a wheeled Turtlebot platform and compare its executable navigation performance against Spot in Table~\ref{tab:robot_performance_main}, highlighting how embodiment and contact dynamics affect trajectory execution.
The benchmark interface is designed to support additional platforms, including quadrupeds, humanoids, and bipedal robots, but these additional embodiments are not experimentally evaluated in this paper.

\subsection{Actions}
\label{sec:app_actions}
NavVerse supports three action types:
\begin{itemize}
  \item \textbf{Waypoint path.} A list of waypoints in the world frame,
  $W=\{(x_i,y_i,\psi_i)\}_{i=1}^N$, where $\psi$ is yaw.
  \item \textbf{Discrete actions.} A discrete primitive action space that approximates waypoint control using fixed step-size motions. Each primitive is converted into a single-step waypoint in the world frame defined in Table~\ref{tab:discrete_actions}.
  \item \textbf{Velocity commands.} A body-frame velocity $(v_x, v_y, \omega)$.
This action type can directly drive the locomotion policy and bypass the PID-based controller.
\end{itemize}

\begin{table}[h]
  \centering
  \caption{Discrete action primitives and their mapped motion in the waypoint interface.}
  \footnotesize
  \renewcommand{\arraystretch}{1.3}
  \setlength{\tabcolsep}{3pt}
  \begin{tabular}{p{0.46\linewidth} p{0.48\linewidth}}
    \toprule
    \textbf{Discrete action} & \textbf{Mapped motion} \\
    \midrule
    \texttt{forward}, \texttt{left}, \texttt{right} &
    $\Delta s = 0.25\,\mathrm{m}$ (indoor); \newline$\Delta s = 1.0\,\mathrm{m}$ (outdoor, indoor-to-outdoor). \\
    \texttt{rotate-left}, \texttt{rotate-right} &
    $\Delta \psi = \pm 30^\circ$. \\
    \texttt{small-rotate-left}, \texttt{small-rotate-right} &
    $\Delta \psi = \pm 15^\circ$. \\
    \bottomrule
  \end{tabular}
  \label{tab:discrete_actions}
\end{table}

\subsection{Action-to-control bridging}
\label{sec:app_bridge}
Waypoints are executed by a shared low-level stack, which runs at $50$ Hz:
\begin{itemize}
  \item \textbf{Path tracking.} Given a waypoint path $\mathbf{W}$ in the world frame, we track the path with lookahead distance $d_{\mathrm{look}}=0.3$ m.
  A waypoint is considered reached when the position error is below $0.03$ m and the yaw error is below $3^\circ$.
  \item \textbf{PID-based controller.} The controller computes body-frame velocity $(v_x, v_y, \omega)$ from the tracking error between the current robot pose and the next lookahead target. We clip it to $|v_x|\le 1.5$ m/s, $|v_y|\le 0.3$ m/s, and $|\omega|\le 1.5$ rad/s.
  \item \textbf{Locomotion policy.} The clipped velocity is passed to a learned locomotion policy with the same structure as the default Isaac Lab locomotion policy.
  We train this policy on flat ground, rough ground, and stairs to support diverse terrain execution.
  \item \textbf{Action execution}
  One discrete action corresponds to approximately 1~s of simulated time, as actions are executed by the locomotion/controller stack until the commanded motion primitive completes.
  The agent can also output waypoint targets, which can be streamed to the controller without blocking on the completion of the previous waypoint execution. This enables non-blocking control and better overlaps perception and planning with low-level execution.

\end{itemize}

\subsection{Episodes}
\label{sec:app_episodes}
An episode starts from an initial robot pose and proceeds step-by-step as the agent outputs actions, until it either succeeds or meets a termination condition.
For ObjNav and PlaceNav, the goal is a category that may have multiple instances in the scene; reaching any valid instance counts as success.
For VLN, the agent is provided a natural-language instruction, and success is evaluated using the same geometric goal criterion as ObjNav and PlaceNav.
In addition to the agent observation, the simulator can output privileged ground-truth signals for logging and training (e.g., SR/SPL, distance to goal, BEV images, goal positions and sizes, and reference trajectories); these signals are not required by the benchmark action interface and are not used by the evaluated policies unless explicitly stated.

\subsection{Metrics}
\label{sec:app_metrics}
NavVerse provides the following metrics: Distance to Goal (DTG), Success Rate (SR), Success weighted by Path Length (SPL), SoftSPL, Navigation Error (NE), and Oracle Success (OS).

Let $P$ be the executed trajectory length accumulated in 3D from the robot base $(x,y,z)$ trajectory.
Let $L$ be the shortest path length computed as the length of the path between the start and the goal.
Let $\mathbf{p}_t$ be the robot base position at time step $t$, and let $T$ be the termination time step.

\begin{itemize}
  \item \textbf{Distance-to-goal.}
  The distance-to-goal at time step $t$ is
  \begin{equation}
  d_t = \max\!\left(0,\, \left\lVert \mathbf{p}_t - \mathbf{p}_{\mathrm{goal}}\right\rVert_2 - r_{\mathrm{goal}}\right),
  \end{equation}
  where $\mathbf{p}_{\mathrm{goal}}$ is the goal center position and
  $r_{\mathrm{goal}}=\tfrac{1}{2}\max(x_{\mathrm{size}}, y_{\mathrm{size}})$ is computed from the goal mesh size in the $x$-$y$ plane.

  \item \textbf{Success.}
  An episode is successful if the final distance-to-goal satisfies
  \begin{equation}
  d_T \le 1.0 + r_{\mathrm{robot}}.
  \end{equation}
  where $r_{\mathrm{robot}}$ is the radius of the robot. We use $r_{\mathrm{robot}}=0.6$~m for Boston Dynamics Spot.

  \item \textbf{Success Rate (SR).}
  \begin{equation}
  \mathrm{SR} = \frac{1}{N}\sum_{k=1}^{N}\mathbb{I}\!\left[d_{T}^{(k)} \le 1.0 + r_{\mathrm{robot}}\right].
  \end{equation}

  \item \textbf{Success weighted by Path Length (SPL).}
  \begin{equation}
  \mathrm{SPL} = \frac{1}{N}\sum_{k=1}^{N}\mathbb{I}\!\left[d_{T}^{(k)} \le 1.0 + r_{\mathrm{robot}}\right]\cdot
  \frac{L^{(k)}}{\max(L^{(k)}, P^{(k)})}.
  \end{equation}

\item \textbf{Soft Success (SS).}
  We use the soft success score $S_p$ to measure partial progress toward the goal:
  \begin{equation}
  S_p = \max\left(0,\, 1 - \frac{d_T}{d_0}\right),
  \end{equation}
  where $d_T$ is the final distance to the goal and $d_0$ is the initial distance.
  We report
  \begin{equation}
  \mathrm{SS} = \frac{1}{N}\sum_{k=1}^{N} S_p^{(k)}.
  \end{equation}

  \item \textbf{SoftSPL.} We define
    \begin{equation}
    \mathrm{SoftSPL}^{(k)}
    =
    S_p^{(k)}
    \frac{L^{(k)}}{\max(L^{(k)},P^{(k)})},
    \end{equation}
    \begin{equation}
    \mathrm{SoftSPL}
    =
    \frac{1}{N}\sum_{k=1}^{N}\mathrm{SoftSPL}^{(k)}.
    \end{equation}
  Here $d_T$ is the distance between the agent's final position and the goal, and $d_0$ is the distance between the start position and the goal.

    \item \textbf{Navigation Error (NE).}
    Navigation error is the distance-to-goal at episode termination:
    \begin{equation}
    \mathrm{NE}
    =
    \frac{1}{N}
    \sum_{k=1}^{N} d_T^{(k)}.
    \end{equation}

  \item \textbf{Oracle Success (OS).}
  Oracle success indicates whether the agent reaches the goal tolerance at any time:
  \begin{equation}
    \mathrm{OS}
    =
    \frac{1}{N}
    \sum_{k=1}^{N}
    \mathbb{I}
    \left[
    \min_{t\in[0,T_k]} d_t^{(k)}
    \le 1.0+r_{\mathrm{robot}}
    \right].
  \end{equation}
  
  \item \textbf{Coverage Efficiency (CE).}
  Let $\mathcal{G}(\tau)$ be the set of unique $1$ m-resolution occupancy-grid cells visited by the executed trajectory $\tau$, and let $P$ be the executed path length.
  Coverage efficiency measures newly explored space per unit travel cost:
  \begin{equation}
  \mathrm{CE} = \frac{|\mathcal{G}(\tau)|}{\max(P,\varepsilon)}.
  \end{equation}
  This metric captures whether an agent explores large open spaces efficiently rather than repeatedly revisiting the same local region.

  \item \textbf{Collision Rate (CR).}
  Let $c_t \in \{0,1\}$ indicate whether the robot has a non-ground collision with obstacle geometry at time step $t$.
  We report the trajectory-level collision ratio:
  \begin{equation}
  \mathrm{CR} = \frac{1}{N}\sum_{k=1}^{N}\mathbb{I}\!\left[\exists t,\; c_t^{(k)}=1\right].
  \end{equation}
  Lower CR indicates safer local motion and better obstacle avoidance.

  \item \textbf{Average Distance to Obstacles (ADO).}
  Let $d^{\mathrm{obs}}_t$ be the distance from the robot base to the nearest obstacle or non-traversable geometry at time step $t$.
  We define
  \begin{equation}
  \mathrm{ADO} = \frac{1}{T}\sum_{t=1}^{T} d^{\mathrm{obs}}_t.
  \end{equation}
  Higher ADO indicates a larger clearance margin during navigation.

  \item \textbf{Navigable Surface Ratio (NSR).}
  Let $n_t \in \{0,1\}$ indicate whether the robot base lies on a valid or kinodynamically preferred navigable surface at time step $t$ according to the scene's terrain annotations.
  We define
  \begin{equation}
  \mathrm{NSR} = \frac{1}{T}\sum_{t=1}^{T} n_t.
  \end{equation}
  Higher NSR indicates that the robot more consistently stays on surfaces suitable for safe traversal.
\end{itemize}

\subsection{Termination labels}
\label{sec:app_termination}
An episode terminates when one of the following conditions is met:
\begin{itemize}
  \item \textbf{Perception error.} The agent calls \texttt{stop} but the goal tolerance is not satisfied, i.e., \texttt{stop} is issued and $d > 1.0 + r_{\mathrm{robot}}$ (equivalently $d>1.6$ m for Spot).
  \item \textbf{Fall.} The angle between the projected gravity vector in the robot frame and the world $z$-axis exceeds $45^\circ$.
  \item \textbf{Timeout.} Simulation time exceeds $300$ seconds.
\end{itemize}

\subsection{System Setup and Runtime}
\label{sec:app_compute}
\begin{itemize}
  \item \textbf{Simulation stack.}
  We use Isaac Sim 5.1.0 with Isaac Lab 2.3.0.
  \item \textbf{Hardware.}
  We use a single RTX~4090 for development and smoke tests, and four NVIDIA L40S GPUs for large-scale evaluation.
  \item \textbf{Physics timestep and rendering.}
  The simulator advances with a fixed physics step of 0.005~s (200~Hz). The rendering throughput is approximately 20~FPS on an RTX 4090.
  \item \textbf{Parallelization.}
  We support parallel evaluation across multiple simulator instances. In our setup, a single NVIDIA L40s (48G VRAM) can run up to three simulations concurrently.
  \item \textbf{Memory footprint.}
  Peak GPU memory usage is around 9~GB for indoor scenes and 13~GB for outdoor and indoor-to-outdoor scenes under the default observation configuration.
\end{itemize}

\begin{figure*}
    \centering
    \includegraphics[width=0.75\linewidth]{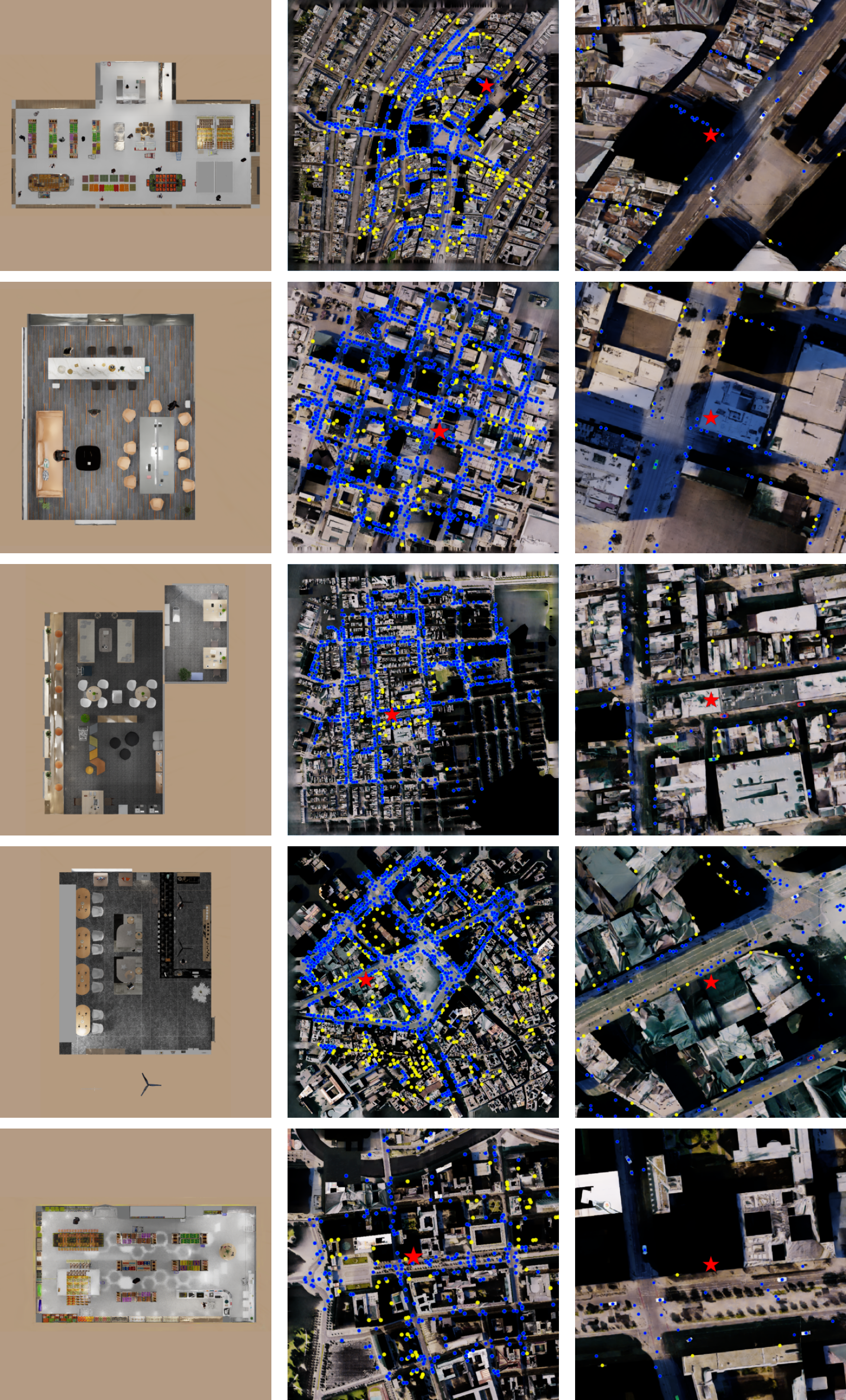}
    \caption{Generated BEV maps for representative indoor, outdoor, and indoor-to-outdoor scenes. Blue points denote object instances and yellow points denote storefront goals. The red star marks the indoor-area anchor used for indoor-to-outdoor scenes and is shown as a city-scale reference in outdoor scenes.}
    \label{fig:bev_maps}
    \vspace{-4mm}
\end{figure*}

\section{Scene Construction}
\label{sec:app_datasets_scenes}
This section provides the implementation details behind the scene-generation summary in Section~\ref{sec:scene_generation}.
\subsection{Scene Sources and Inventory}
\label{sec:app_scene_inventory}
\begin{itemize}
  \item \textbf{Indoor scenes.}
  We use 100 indoor scenes from \textit{GRScenes} in GRUtopia~\cite{wang2024grutopia}, covering apartments, hospitals, cafes, supermarkets, offices, and other everyday environments.
  The split uses 90 indoor scenes for training and 10 for evaluation.
  \item \textbf{Outdoor scenes.}
  We select 50 cities from the \textit{Virtual Community} dataset~\cite{zhou_virtual_nodate} to construct our benchmark, with 40 scenes for training and 10 for evaluation.
  Each scene covers a 1\,km$\times$1\,km area.
  Terrain across outdoor scenes is segmented into five classes: buildings, sidewalks, roadways, water bodies, and green areas.
  On average, each city contains 546 buildings, 36.4\,km of road network, and 199{,}530\,m$^2$ of navigable area.
  \item \textbf{Indoor-to-outdoor scenes.}
  The benchmark includes 50 indoor-to-outdoor scenes created by embedding indoor layouts into outdoor cities through door-to-facade connections.
  Indoor-to-outdoor scenes follow the same scene-disjoint split protocol as the outdoor set.
  \item \textbf{Object and storefront inventory.}
  We curate 121 Objaverse~\cite{deitke2023objaverse} object categories suitable for urban environments, such as vehicles, chairs, and trash cans.
  Outdoor scenes populate lanes with up to 200 cars and sidewalks with up to 1{,}000 objects per scene.
  Indoor-to-outdoor scenes sample objects and cars within a 50\,m radius of the indoor anchor to control episode difficulty.
  We additionally add 30 contextual-object groups and 101 generated storefront textures, as detailed in Sections~\ref{sec:app_contextual_objects} and~\ref{sec:app_storefronts}.
  \item \textbf{BEV map examples.}
  Figure~\ref{fig:bev_maps} shows representative BEV maps for indoor, outdoor, and indoor-to-outdoor scenes.
  Indoor scenes cover diverse layout types and contain rich object-category compositions.
  Outdoor scenes span large urban areas, with object instances and storefront goals distributed according to the underlying map structure.
  In indoor-to-outdoor scenes, many objects and storefronts are placed around the indoor-area anchor, increasing local goal density near the indoor region and improving task diversity for connected episodes.
\end{itemize}
\subsection{Indoor Scene Processing}
\label{sec:app_indoor_processing}
\begin{itemize}
  \item \textbf{Geometry preprocessing.}
  We develop an automated mesh-optimization pipeline to reduce scene complexity and accelerate simulation and navigation.
  The pipeline applies three geometry operations in Blender: decimation to reduce face count, dissolve to remove redundant edges and vertices, and remeshing to regularize the mesh while preserving the overall structure.
  The optimized meshes keep the original textures and materials.
  Table~\ref{tab:mesh_decimation} summarizes the reduction in mesh complexity.
  This optimization substantially speeds up NavMesh construction, reducing build time from 705\,ms to 107\,ms, and also reduces scene loading time.
    \begin{table}[h]
    \centering
    \caption{Average face count before and after decimation.}
    \label{tab:mesh_decimation}
    \begin{tabular}{lccc}
    \hline
    Scene Type & Raw Faces & Decimated Faces & Decimation (\%) \\
    \hline
    GRScenes &
    21,774,867 &
    2,197,731 &
    89.91 \\
    \hline
    \end{tabular}
    \end{table}
\end{itemize}
\subsection{Outdoor Terrain and Road Modeling}
\label{sec:app_outdoor_terrain}
\begin{itemize}
  \item \textbf{Terrain generation.}
  We start from the imported city terrain mesh and densify it to provide sufficient geometric resolution for road modeling and ground-level execution.
  This is done through a lightweight cleanup and subdivision procedure that increases vertex density while preserving the overall terrain shape.
  \item \textbf{Road depression.}
  We extract vehicle road centerlines and widths from OSM, using tagged widths when available and otherwise deriving widths from lane count and road type.
  We convert these road attributes into road footprint polygons.
  We imprint the road footprints onto the terrain mesh and apply a uniform depression depth of $0.2$\,m inside road regions.
  This produces physically traversable roadbeds that align with the map geometry.
  \item \textbf{Surface semantics and materials.}
  We label terrain regions into classes such as sidewalks, roadways, water bodies, and green areas.
  We assign class-specific appearance and physics materials so that policies experience different friction and contact behavior across surfaces.
\end{itemize}
\subsection{Contextual Objects}
\label{sec:app_contextual_objects}
We enrich outdoor and indoor-to-outdoor scenes with contextual object groups to increase visual realism, provide local semantic cues, and diversify object-level goals.
Unlike random object scattering, contextual placement is conditioned on the semantic type of each map region.
For each region, we sample only object groups that fit the local context, such as bike racks and bicycles in bike-parking areas or bus shelters, briefcases, and umbrellas near bus stops.
Contextual objects are instantiated from a curated library of 30 \emph{contextual-object groups}.
Each group consists of a small set of assets that co-occur in realistic activities, and objects may be stacked based on their sizes and support surfaces, such as a bag on a bench or toys on a table.
Table~\ref{tab:contextual_groups} shows representative examples of the group definitions.
To avoid excessive clutter, the placement of a new group is constrained to be at least $8.0\,\mathrm{m}$ away from any existing contextual group.
\begin{table}[t]
  \centering
  \caption{Example contextual object groups. In the benchmark, each group corresponds to a set of USD assets.}
  \small
  \setlength{\tabcolsep}{6pt}
  \begin{tabular}{l l}
    \toprule
    \textbf{Group name} & \textbf{Assets} \\
    \midrule
    feeding\_cat & cat, rice\_bowl, nursing\_bottle \\
    park\_relaxing & park\_bench, notebook, coffee\_cup \\
    bus\_stop\_wait & bus\_stop, briefcase, umbrella \\
    construction\_repair & ladder, cargo\_pallet \\
    sports\_gear & soccer\_ball, sports\_jacket, shoes \\
    \bottomrule
  \end{tabular}
  \label{tab:contextual_groups}
\end{table}
\subsection{Storefronts}
\label{sec:app_storefronts}
We create generated storefront assets to enrich outdoor building facades and provide visually distinctive goal locations for PlaceNav and language-guided tasks.
We generate 101 storefront textures covering place categories such as restaurants, cafes, and banks, together with corresponding relative depth maps using \textit{Gemini 3 Pro Image}~\cite{nano_banana_pro}.
The generated assets are manually inspected for visual plausibility and semantic consistency.
Each accepted storefront is converted into a 3D mesh by depth-based extrusion in Blender.
We place storefront meshes at OSM-specified store locations by ray-casting to the nearest building facade, favoring vertical exterior faces.
We orient each storefront toward the closest road, scale it to a target height, and align its base to the terrain.
We carve a matching opening in the facade, shift laterally to resolve corner or overlap conflicts, and accept a placement only if the cut succeeds, the storefront faces the road, and it remains unoccluded by surrounding geometry.
\subsection{Indoor-to-Outdoor Scene Assembly}
\label{sec:app_indoor_to_outdoor_assembly}
We construct indoor-to-outdoor scenes by embedding indoor layouts into buildings in the outdoor city environment.
For each scene, we first search for a suitable road-facing building facade with enough space for an exit and nearby traversable street or sidewalk area.
We create a facade opening to serve as the outdoor exit.
We then place a compatible indoor layout behind the selected building facade.
We align the indoor entrance with the facade opening and adjust the entrance height to match the local outdoor terrain.
We remove blocking door, wall, or facade meshes around the connection, producing a traversable passage between the interior and the street.
This door-to-facade connection allows the agent to transition smoothly between the interior and the street without teleportation or viewpoint switching.
The indoor and outdoor components are executed in the same Isaac Sim scene, so the robot uses the same physics, sensing, and control stack across the full trajectory.
The resulting indoor-to-outdoor scenes support episodes that start inside an indoor room, pass through interior corridors, exit into the city, and continue toward outdoor goals.
These episodes introduce abrupt changes in visual context, lighting, environmental structure, and spatial scale, enabling evaluation of whether agents can preserve the goal and adapt their search behavior after crossing from indoor layouts into outdoor space.
\subsection{NavMesh Generation}
\label{sec:app_navmesh}
We generate navigation meshes with the C++ \texttt{RecastNavigation} library and provide a lightweight Python wrapper for NavMesh building and query APIs.
\begin{itemize}
  \item \textbf{Adaptive settings.}
  We use two NavMesh presets to support scenes with different scales.
  Indoor scenes use a finer voxelization with smaller cell size and height to capture narrow traversable regions.
  Outdoor scenes use a coarser resolution for efficiency.
  Table~\ref{tab:navmesh_params} summarizes the NavMesh build parameters used in this work.
  \item \textbf{Episode sampling and reference paths.}
  We use the NavMesh to sample feasible start positions and to generate reference paths from the start to each goal instance.
  These paths respect traversability constraints encoded by the NavMesh build parameters, including maximum walkable slope and maximum climb, and therefore reflect what the robot can physically traverse.
\end{itemize}
\begin{table}[t]
  \centering
  \caption{NavMesh build parameters used for indoor and outdoor scenes. Indoor scenes use finer discretization to better represent narrow traversable regions. All length parameters are in meters.}
  \small
  \setlength{\tabcolsep}{6pt}
  \begin{tabular}{lcc}
    \toprule
    \textbf{Parameter} & \textbf{Indoor} & \textbf{Outdoor} \\
    \midrule
    agent\_height & 0.60 & 0.60 \\
    agent\_radius & 0.60 & 0.60 \\
    agent\_max\_climb & 0.20 & 0.30 \\
    agent\_max\_slope (deg) & 45.0 & 45.0 \\
    cell\_size & 0.05 & 0.10 \\
    cell\_height & 0.05 & 0.10 \\
    region\_min\_size & 8.0 & 8.0 \\
    region\_merge\_size & 0.0 & 0.0 \\
    edge\_max\_error & 1.5 & 1.5 \\
    edge\_max\_len & 0.0 & 0.0 \\
    detail\_sample\_dist & 1.0 & 1.0 \\
    detail\_sample\_max\_error & 0.0 & 0.0 \\
    verts\_per\_poly & 6.0 & 6.0 \\
    \bottomrule
  \end{tabular}
  \label{tab:navmesh_params}
\end{table}

\section{Task Generation}
\label{sec:app_task_generation}

\subsection{Episode Sampling}
\label{sec:app_episode_sampling}
Episodes are generated by sampling a starting position and a goal category, then constructing a set of valid goal instances and their oracle reference paths.

\begin{itemize}
  \item \textbf{Goal discovery and instances.} We traverse scene prims and group them into goal categories with scene-specific matching rules (e.g., regular expressions over prim names or prim paths).
  \item \textbf{NavMesh-based start sampling and reference paths.} We sample start positions from the traversable region of the scene NavMesh and compute a path from the sampled start position to each candidate goal instance center. For each scene, we sample goal categories uniformly. We retain instances only if the path lies within a prescribed range ($5.0\,\mathrm{m} \le L \le 50.0\,\mathrm{m}$). 
  \item \textbf{Episode initialization.} If at least one instance is retained, we form an episode. The start yaw is initialized to face the next waypoint on the closest-instance reference path, plus a random perturbation sampled uniformly from $[-30^\circ, 30^\circ]$.
  \item \textbf{Indoor-to-outdoor task sampling.} For indoor-to-outdoor scenes, the start position is constrained to lie inside the embedded indoor region. Candidate outdoor goal instances are filtered to a bounded neighborhood around the indoor region to avoid excessively long or irrelevant indoor-to-outdoor tasks.
\end{itemize}

\subsection{Online Verification and Quality Control}
\label{sec:app_episode_verification}
After offline sampling, we verify each candidate episode by executing a rollout that tracks the reference waypoints under full physics. We log observations and poses along the rollout for traceability. An episode is accepted if it fulfill all of these conditions:

\begin{itemize}
  \item \textbf{Oracle feasibility.} A candidate episode must achieve oracle success with respect to at least one goal instance at some time during the trajectory.
  \item \textbf{Episode duration.} The rollout duration must exceed $5.0$~s  in simulation time to avoid degenerate episodes, and must be no more than $100$~s during task generation to avoid overly difficult episodes and keep dataset construction efficient.
  \item \textbf{Traveled distance.} The executed trajectory length during verification must exceed $5.0$~m but less than $50.0$~m.
  \item \textbf{No termination.} Episodes that terminate due to failure conditions (e.g., out-of-bounds, fall, stuck) are rejected.
\end{itemize}

\subsection{VLN Instruction Generation and Filtering}
\label{sec:app_vln_generation}
For VLN, we attach a natural-language instruction to each verified episode. Instruction generation proceeds in two stages, and episodes that fail automatic alignment or human checks are excluded from VLN evaluation.

\begin{itemize}
  \item \textbf{Template instructions.} We generate a template instruction by converting a simplified reference path into a compact sequence of turn-and-move directives, followed by an arrival statement.
  \item \textbf{Trajectory-to-frame alignment.} Using the recorded rollout poses, we align directives to frames by matching simplified waypoints to nearby executed poses.
  \item \textbf{VLM refinement.} We refine the template using a vision-language model by providing a sparse set of keyframes sampled along the rollout (similar to Figure~\ref{fig:gttraj_examples}) and requesting a concise instruction that emphasizes visually distinctive landmarks while avoiding exact distance values. The prompt details are provided in Box~\ref{box:vlm_prompt}.
  \item \textbf{Human filtering.} We additionally perform human checks to remove episodes with poor visual quality (e.g., bad lighting) or other issues (e.g., wrong VLM instruction) that hinder instruction-following evaluation.
\end{itemize}

\begin{tcolorbox}[
  title={VLN instruction refinement prompt},
  label={box:vlm_prompt},
  breakable,
  colback=gray!5,
  colframe=gray!40,
  colbacktitle=gray!60,
  boxrule=0.4pt,
  arc=2pt,
  left=4pt,right=4pt,top=4pt,bottom=4pt
]
\begin{lstlisting}[style=promptstyle]
Describe the trajectory of the robot in the images. Improve the given instruction to:
1) include unique landmarks you see in the images for guidance, describe color, shape, etc. briefly. Only mention landmarks that can be clearly observed.
2) make the expression more diverse.
3) replace the exact distance value with more general description.
4) do not mention (img x) in the output.
5) in some cases, if you don't directly see the target in the image, just say "you will arrive at [target name]".
6) do not mention the red lines in the images, those are pointing to the waypoints.
7) if the robot is following the road, mention it in the output.
8) make the instruction compact, only include the key landmarks.
9) return the output only, no other text.
10) make it within 100 words.

Improve this instruction: {PROMPT_INSTRUCTION}

Example output: Go forward until you see the red building, then turn right and go until you see the blue car. You should be able to see the target mailbox on your right.
\end{lstlisting}
\end{tcolorbox}

\section{Benchmark Statistics}
\label{sec:app_benchmark_statistics}

\subsection{Episode Overview}
\label{sec:app_stats_counts}

\begin{table*}[t]
  \caption{\textbf{Episode counts by split and task.}
  We use Train and Eval splits with 10K episodes in total.
  The Eval split follows the quality-filtered evaluation set, so task and scene counts are not perfectly balanced.}
  \label{tab:episode_counts}
  \centering
  \footnotesize
  \setlength{\tabcolsep}{3pt}
  \renewcommand{\arraystretch}{1.05}
  \resizebox{\textwidth}{!}{%
  \begin{tabular}{lccc ccc ccc r}
    \toprule
    & \multicolumn{3}{c}{\textbf{ObjNav}}
    & \multicolumn{3}{c}{\textbf{PlaceNav}}
    & \multicolumn{3}{c}{\textbf{VLN}}
    & \multirow{2}{*}{\textbf{Total}} \\
    \cmidrule(lr){2-4}\cmidrule(lr){5-7}\cmidrule(lr){8-10}
    \textbf{Split}
    & \textbf{Indoor} & \textbf{Outdoor} & \textbf{Indoor-to-Outdoor}
    & \textbf{Indoor} & \textbf{Outdoor} & \textbf{Indoor-to-Outdoor}
    & \textbf{Indoor} & \textbf{Outdoor} & \textbf{Indoor-to-Outdoor}
    & \\
    \midrule
    Train & 2{,}200 & 700 & 800 & 0 & 1{,}450 & 1{,}400 & 850 & 850 & 850 & 9{,}100 \\
    Eval  & 150 & 82 & 95 & 0 & 68 & 55 & 150 & 150 & 150 & 900 \\
    \midrule
    Total & 2{,}350 & 782 & 895 & 0 & 1{,}518 & 1{,}455 & 1{,}000 & 1{,}000 & 1{,}000 & 10{,}000 \\
    \bottomrule
  \end{tabular}%
  }
\end{table*}

\begin{itemize}
  \item \textbf{Split Protocol.}
  We construct train and eval splits at the scene level, and scenes do not overlap across splits. This prevents layout leakage and ensures that policies must generalize to novel environments rather than memorizing scene-specific geometry and appearance.
  \item \textbf{Counts.}
The released dataset contains 10{,}000 episodes, including 4{,}027 ObjNav episodes, 2{,}973 PlaceNav episodes, and 3{,}000 VLN episodes. We separate them into two splits, Train and Eval. Both splits are obtained after episode-quality filtering, so the task and scene counts are not exactly balanced. In particular, the dataset contains more indoor ObjNav episodes, while outdoor and indoor-to-outdoor scenes contain fewer ObjNav episodes and relatively more PlaceNav episodes. VLN episodes are generated from the same underlying navigation tasks by producing language instructions along the reference path. Some candidates are removed because they contain too few distinctive landmarks for instruction following, or because the VLM-based instruction refinement fails quality checks, resulting in 3{,}000 VLN episodes in total.
\end{itemize}

\subsection{Goal Distribution}
\label{sec:app_goal_dist}

\begin{figure}[t]
  \centering
  \includegraphics[width=0.72\linewidth]{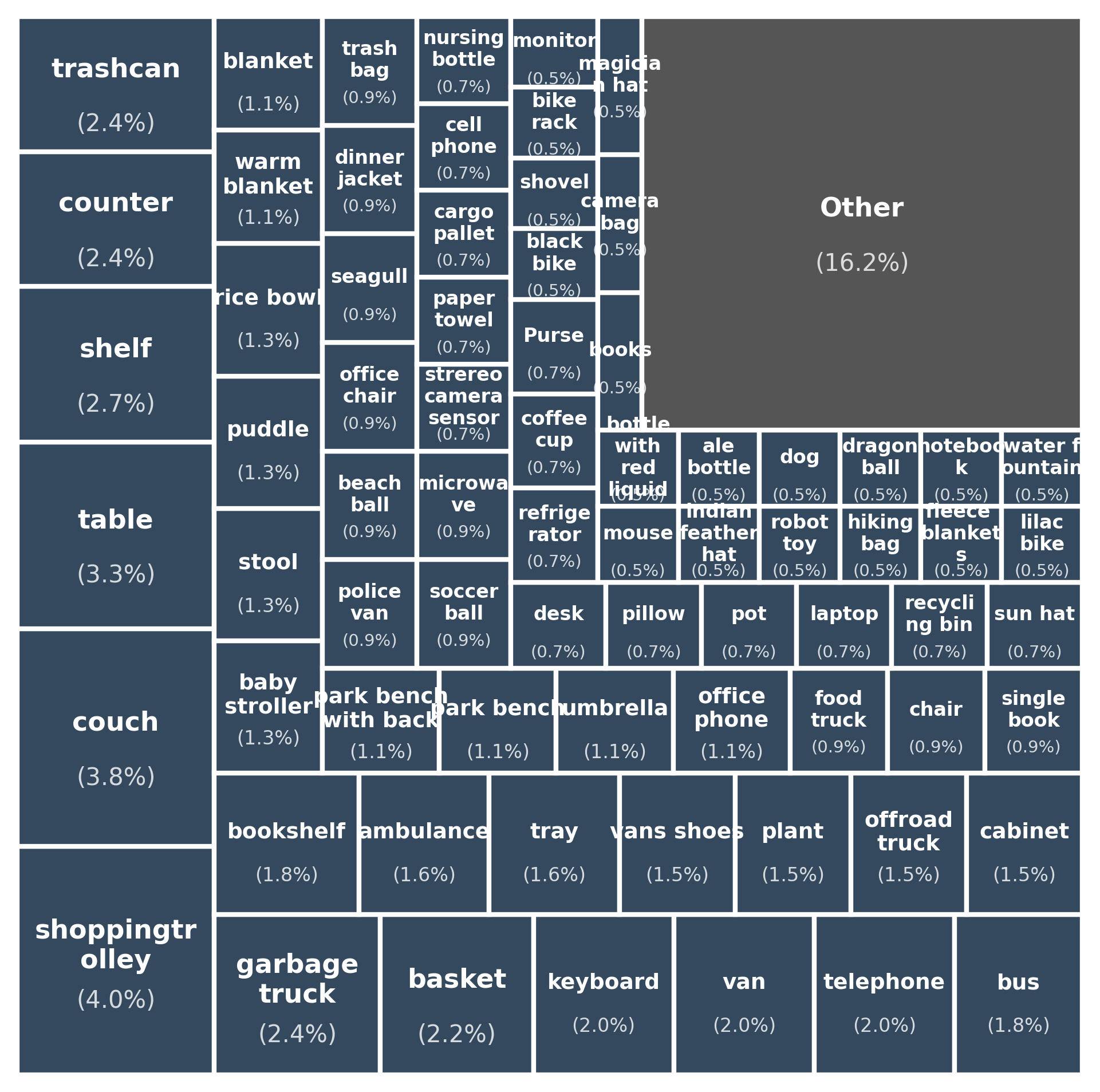}
  \caption{ObjNav target category distribution.}
  \label{fig:objnav_treemap}

  \vspace{0.1cm}
  
  \includegraphics[width=0.72\linewidth]{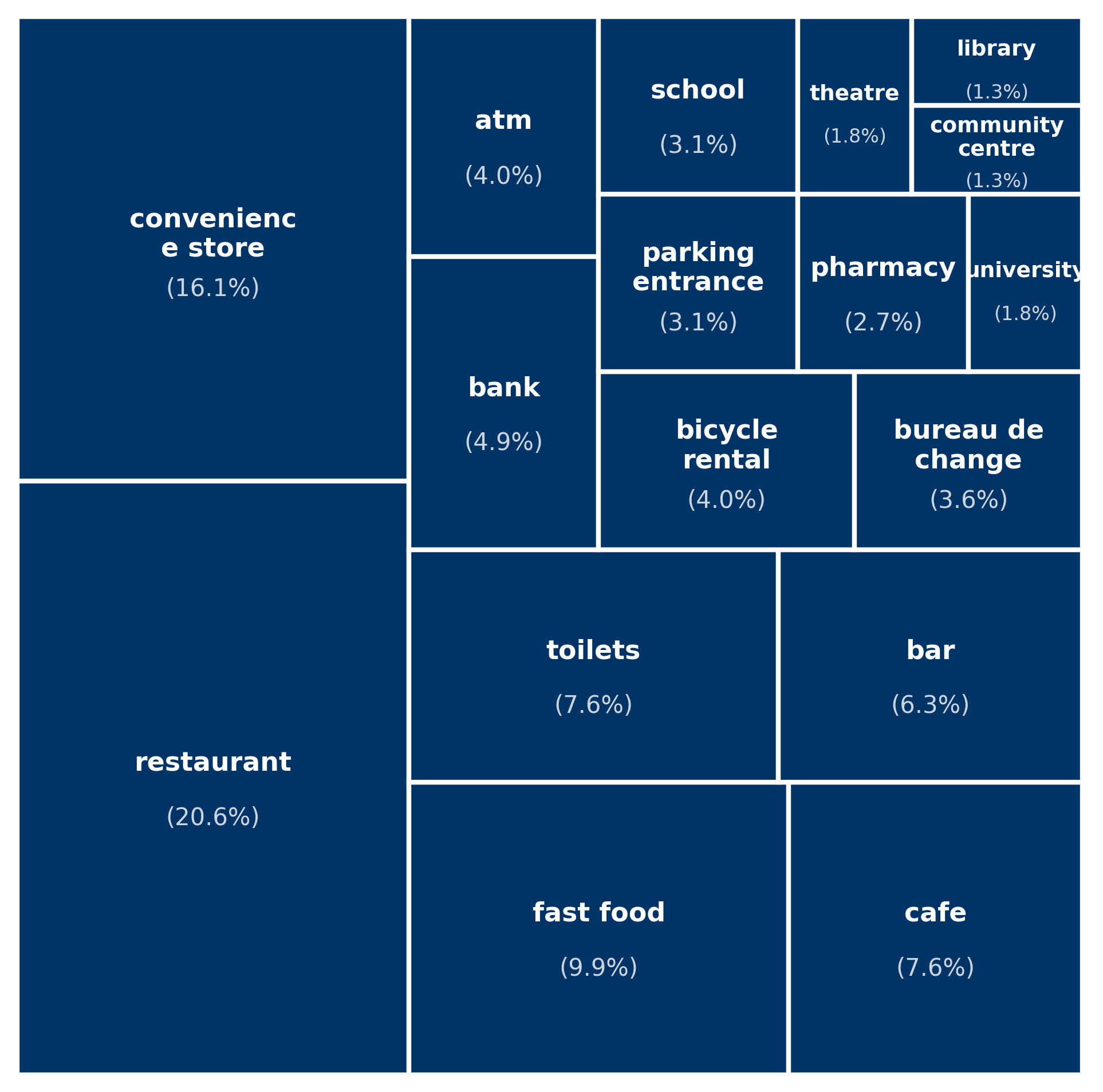}
  \caption{PlaceNav storefront category distribution.}
  \label{fig:placenav_treemap}
\end{figure}
\begin{itemize}
  \item \textbf{ObjNav.}
  ObjNav includes 136 goal categories. Figure~\ref{fig:objnav_treemap} summarizes the target distribution. Several categories appear frequently because they correspond to common indoor objects that occur repeatedly across interior layouts. In contrast, outdoor scenes contribute a larger variety of categories, producing a more diverse and long-tailed distribution. This long tail increases the difficulty of perception, since many targets have limited exposure compared to the most frequent indoor categories.
\end{itemize}

\begin{itemize}
  \item \textbf{PlaceNav.}
  PlaceNav includes 17 storefront goal categories. Figure~\ref{fig:placenav_treemap} shows that the distribution concentrates on common place types (e.g., restaurants, convenience stores, cafes), which aligns with real-world urban statistics and OSM-derived semantics. This indicates that the benchmark captures realistic place frequencies while still covering a broad set of storefront types for evaluation.
\end{itemize}

\subsection{Instruction Vocabulary}
\label{sec:app_vln_vocab}

\begin{figure}[th]
  \centering
  \includegraphics[width=0.76\linewidth]{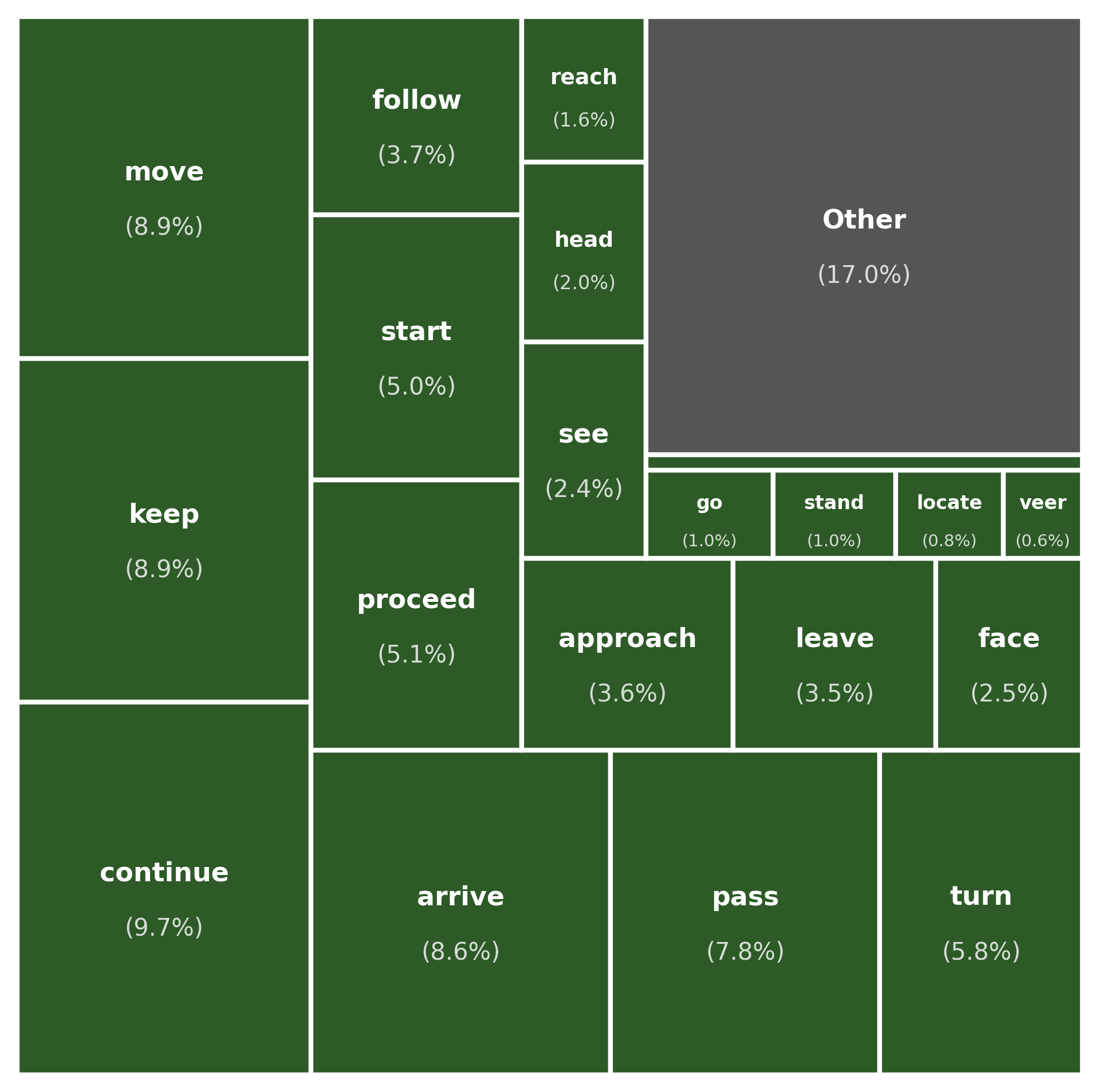}
  \includegraphics[width=0.76\linewidth]{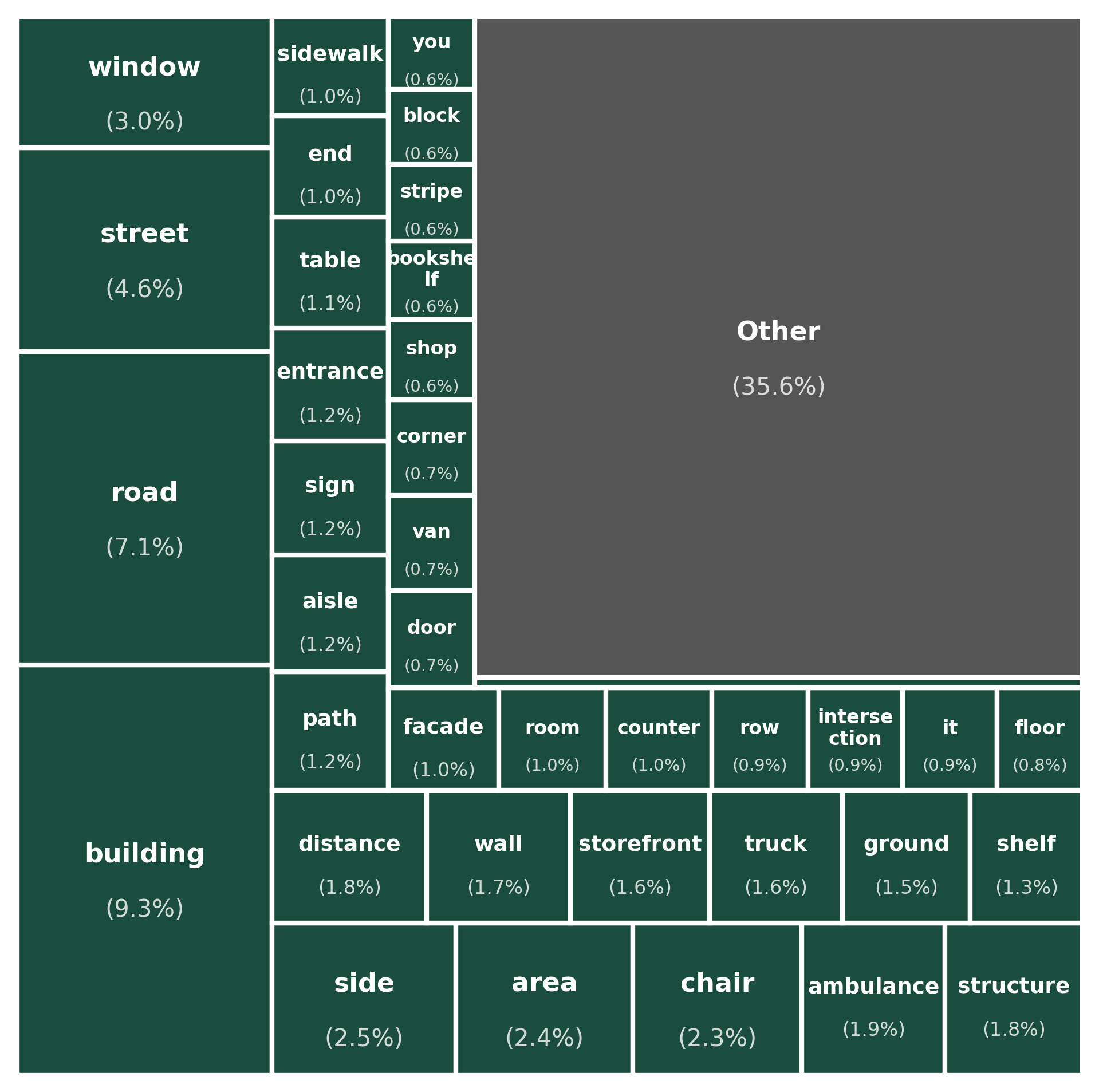}
  \caption{Treemaps of the most frequent tokens in VLN instructions. Top: action verbs. Bottom: landmark/object nouns.}
  \label{fig:stats_vln_token}
\end{figure}

\begin{itemize}
  \item \textbf{Action verbs.}
  Figure~\ref{fig:stats_vln_token} (top) summarizes the verb distribution in VLN instructions. High-frequency verbs emphasize forward progress and route following (e.g., \emph{continue}, \emph{move}, \emph{arrive}, \emph{pass}, \emph{turn}), indicating that most instructions describe incremental motion primitives and local heading changes. The presence of mid-frequency verbs such as \emph{follow}, \emph{approach}, and \emph{reach} suggests that instructions also include goal-directed segments and landmark-conditioned transitions, while the long tail (\textit{Other}) reflects additional fine-grained phrasing that appears less often.
  \item \textbf{Landmark and object nouns.}
  Figure~\ref{fig:stats_vln_token} (bottom) shows the noun distribution used for describing landmarks and targets. Frequent nouns (e.g., \emph{building}, \emph{road}, \emph{street}, \emph{window}) capture global outdoor structure and traversable space, while indoor-oriented nouns (e.g., \emph{table}, \emph{door}, \emph{aisle}) provide local cues for fine navigation and goal localization. The broad coverage across both structural landmarks and object-level references indicates that the VLN set supports multi-scale grounding.
\end{itemize}

\section{Feasibility Analysis}
\label{sec:app_feasibility}

\begin{table*}[t]
\caption{Feasibility analysis of SGImagineNav~\cite{hu_imaginative_2025} under three GT information settings. GT Trajectory calibrates physical executability, while GT Goal helps isolate the transition difficulty beyond target localization.}
\label{tab:feasibility}
\small
\centering
\setlength{\tabcolsep}{2.5pt}
\renewcommand{\arraystretch}{0.92}
\begin{tabular}{l l c c c c c c c c}
\toprule
\multirow{2}{*}{Task} & \multirow{2}{*}{Setting} &
\multicolumn{2}{c}{\textbf{All Scenes}} &
\multicolumn{2}{c}{\textbf{Indoor}} &
\multicolumn{2}{c}{\textbf{Outdoor}} &
\multicolumn{2}{c}{\textbf{Indoor-to-Outdoor}} \\
\cmidrule(lr){3-4} \cmidrule(lr){5-6} \cmidrule(lr){7-8} \cmidrule(lr){9-10}
&& SR $\uparrow$ & SPL $\uparrow$ & SR $\uparrow$ & SPL $\uparrow$ & SR $\uparrow$ & SPL $\uparrow$ & SR $\uparrow$ & SPL $\uparrow$ \\
\midrule
\multirow{3}{*}{ObjNav}
& No GT & 6.42 & 0.61 & 10.00 & 0.73 & 2.44 & 0.21 & 4.21 & 0.74 \\
& GT Goal & 27.83 & 16.46 & 22.22 & 8.57 & 56.63 & 42.20 & 12.00 & 6.46 \\
& GT Trajectory & 100.00 & 90.16 & 100.00 & 83.37 & 100.00 & 96.16 & 100.00 & 95.70 \\
\midrule
\multirow{3}{*}{PlaceNav}
& No GT & 4.88 & 0.64 & - & - & 5.88 & 0.90 & 3.64 & 0.32 \\
& GT Goal & 46.58 & 27.11 & - & - & 48.53 & 28.37 & 20.00 & 9.89 \\
& GT Trajectory & 100.00 & 96.33 & - & - & 100.00 & 96.39 & 100.00 & 95.49 \\
\bottomrule
\end{tabular}
\end{table*}

We conduct a feasibility study with SGImagineNav~\cite{hu_imaginative_2025} to isolate which components limit end-to-end performance. We evaluate three settings: \textbf{No GT} (no privileged goal or trajectory), \textbf{GT Goal} (oracle goal location provided to the planner), and \textbf{GT Trajectory} (execution driven by an oracle reference path/waypoints).

Table~\ref{tab:feasibility} reports ObjNav and PlaceNav results across scene types. GT Goal substantially improves SR/SPL (e.g., ObjNav all-scenes SR from 6.42\% to 27.83\%), while GT Trajectory achieves 100\% SR in all domains, indicating the stack can succeed when the commanded route is correct. SPL remains below 100\% under GT Trajectory (e.g., 90.16\% on all-scenes ObjNav) due to tracking error from the PID controller and the learned locomotion policy, which introduces small deviations under physics. The large gap between No GT and GT Goal suggests that traversability under physics, narrow indoor passages, and outdoor terrain variability are major challenges even when the goal is known. Indoor-to-outdoor episodes remain difficult without oracle information: No GT achieves only 4.21\% SR on ObjNav and 3.64\% on PlaceNav, while GT Goal and GT Trajectory substantially improve success.

\begin{figure*}[b]
  \centering
  \includegraphics[width=0.75\linewidth]{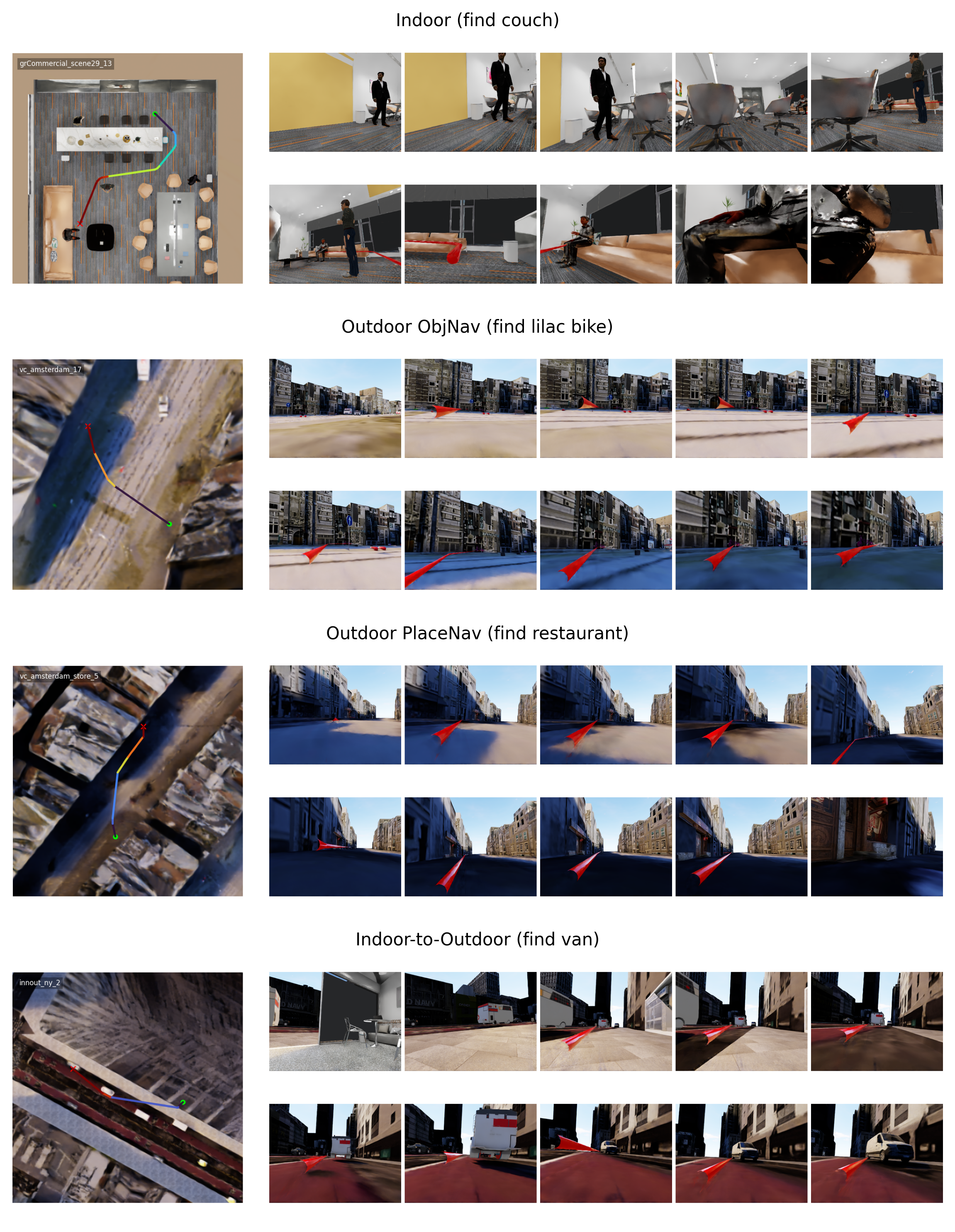}
  \caption{\textbf{GT Trajectory} qualitative examples. Each row shows one episode: the left panel is a BEV overview with the oracle reference path, and the right panels are egocentric RGB frames. The start is marked by a yellow dot and the goal by a red cross in the BEV, and the red overlay in RGB frames indicates the commanded oracle reference path.}
\label{fig:gttraj_examples}
\end{figure*}

Figure~\ref{fig:gttraj_examples} provides qualitative diagnostics for the \textbf{GT Trajectory} setting by pairing a BEV overview with egocentric rollout frames from the same episode. The BEV panels show the oracle route, and the RGB frames illustrate the observations encountered during execution across indoor, outdoor, and indoor-to-outdoor scenes, including typical landmarks, viewpoint changes, and visibility variations along the path.

\end{document}